# Fractal AI

## A Fragile Theory of Intelligence

Sergio Hernández Cerezo
Guillem Duran Ballester

**FrAGIle**

## BOOK #1

### “Forward Thinking”

Version: V4.1

**Main researchers**
Sergio Hernández Cerezo (@EntropyFarmer)
Guillem Duran Ballester (@Miau_DB)

**Special thanks**
Researchers' families for suffering us in all the 'eureka' moments.
HCSoft and Source{d} for their unconditional support.

**Reviewers**
Eiso Kant (@EisoKant), CEO at source{d}
José María Amigó García (Elche University)
Roshawn Terrell (@RoshawnTerrell)
Juan G. Cruz Ayoroa
Jesús P. Nieto (@HedgeFair)
Aidan Rocke (@AidanRocke)
Spiros Baxevanakis(@SpirosBax)
Brian Njenga
Trevor Gower
Anton Osika (@AntonOsika)
Doug Mayhew

# 1 - Introduction

---

*"For instance, on the planet Earth, man had always assumed that he was more intelligent than dolphins because he had achieved so much—the wheel, New York, wars and so on—whilst all the dolphins had ever done was muck about in the water having a good time. But conversely, the dolphins had always believed that they were far more intelligent than man—for precisely the same reasons."*

---

***Douglas Adams, The Hitchhiker's Guide to the Galaxy***

One of the big obstacles in the field of artificial intelligence is not having a definition of intelligence based on solid mathematical and physical principles that could inspire the design and implementations of efficient intelligent algorithms.

For instance, consider the most widely accepted definition of intelligence, signed by 52 specialist on the field [2]:

> *"A very general mental capability that, among other things, involves the ability to reason, plan, solve problems, think abstractly, comprehend complex ideas, learn quickly and learn from experience. It is not merely book learning, a narrow academic skill, or test-taking smarts. Rather, it reflects a broader and deeper capability for comprehending our surroundings..."*

A more recent definition [3] provided by Shane Legg, chief scientist of Deep Mind, and Marcus Hutter, founder of AIXI, is the following:

> *"Intelligence measures an agent's ability to achieve goals in a wide range of environments."*

Although there are many other definitions of intelligence, they are too fuzzy to help us develop a theory of intelligent behaviour or give us an insight on how a general, computable and efficient algorithm for generating intelligent behaviour should look like.

This document is an effort to present such a definition based on entropic principles deeply inspired by the concept of "Causal Entropic Forces" introduced by Alexander Wissner-Gross in 2013 [1] and to propose a generic implementation of those principles.

## 1.1 - The playground of intelligence

As a first attempt in defining intelligence, we could say that intelligence works by taking decisions that directly affect the degrees of freedom of a system in such a way that its future evolution is biased toward rewarding futures.

### 1.1.1 - Cart-pole example

Let's suppose we are controlling a cart-pole that can move left or right by pushing one of the two available buttons. Our goal is to keep the pole standing up, so the ball at the tip of the cart-pole must be as high as possible.

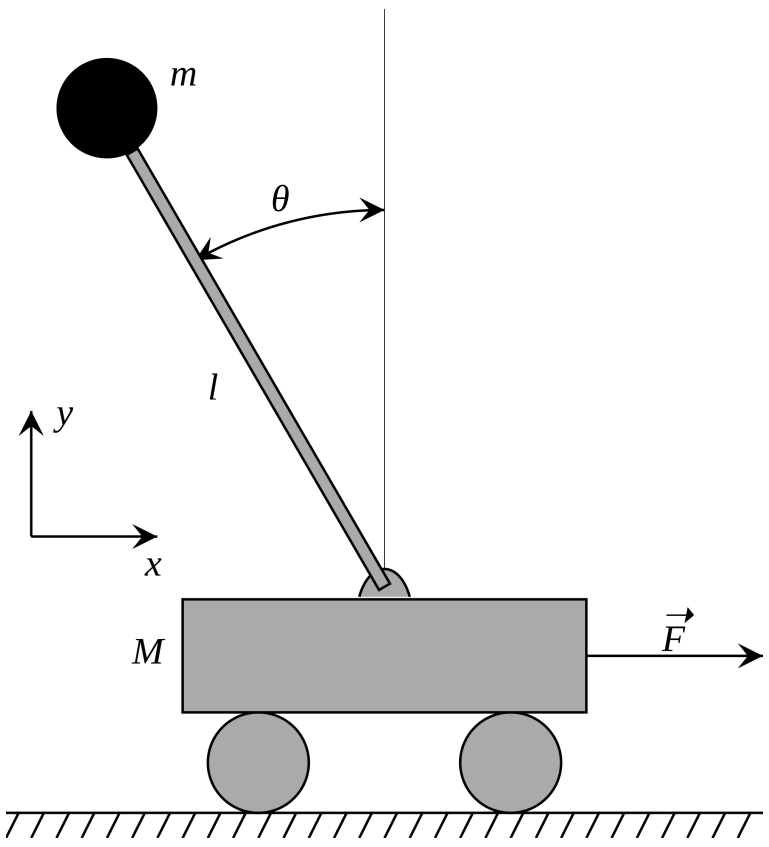

In that case, the intelligence can affect the system evolution by pressing one of the two buttons at any time. The ultimate goal of the intelligence is then to continuously choose the actions that keeps the ball as high as possible.

In this example, the action-space is discrete, but it could also be continuous: instead of having two buttons to choose from, we may have a "joystick" we can push, so our available actions are real numbers in the range [-1, 1]. The simulation of the system can also be more or less deterministic, and the goals could be a combination of several sub-goals. None of this would change the problem except that they would require more computation.

### 1.1.2 - General strategy

When the intelligence is asked to choose between a discrete set of actions $\{a_i\}$ it will internally score them accordingly to some metrics and then output an "intelligent decision" as being the action with the highest score or, in the continuous case, the average of a number of actions weighted by their normalised scores.

$$\text{Decision} = \sum(a_i * \text{Score}(a_i)) / \sum(\text{Score}(a_i))$$

#### 1.1.2.1 - Forward vs Backward intelligence

> *"You can know the past, but not control it. You can control the future, but have not knowledge of it."*
>
> **Claude Shannon**

There are two main strategies used in an intelligent decision making process:

On one hand, we can use information from past events, along with the decisions that were taken and their corresponding outcomes, and eventually learn from that information, to influence future decisions. We will refer to this as 'backward-thinking', as we will only base our decisions on events from the past.

Such a backward-thinking process could in fact learn to predict the best action given an initial state, but it could also learn to predict the next state of the system, as it has access to pairs of initial states, actions, and final states. Predicting the final state from the initial one is equivalent to being able to internally simulate the system for relatively long periods by simulating one state after the other in small jumps.

At some point, evolution began to develop agents that were able to project their actual state into the future with relative accuracy, enabling it to ponder about its available actions in terms not only of its past experiences, but by predicting the foreseeable future each action leads to and its consequences.

This 'forward-thinking' process is a planning algorithm that will make better decisions as the simulation gets better, as opposed to learning-based strategies of back-thinking that depend on the accumulation of past experiences to improve.

Both strategies are complementary. You need to learn how to simulate the system before you can start thinking forward, while forward-thinking can be used to detect and develop better decisions for situations where only repeating past strategies, is no longer viable.

This document will focus on forward-thinking, or the ability to make near-perfect intelligent decisions based on a near-perfect simulation of the system, without the need of any previous learning. This is present in the human cognitive processes and, in current AI methods, it is referred to as "planning algorithms".

#### 1.1.2.2 - Scoring actions

> *"No sensible decision can be made any longer without taking into account not only the world as it is, but the world as it will be."*
>
> **Isaac Asimov**

So what is the basic idea behind "scoring" an action? Imagine the cart holding the pole is in the situation shown in the image:

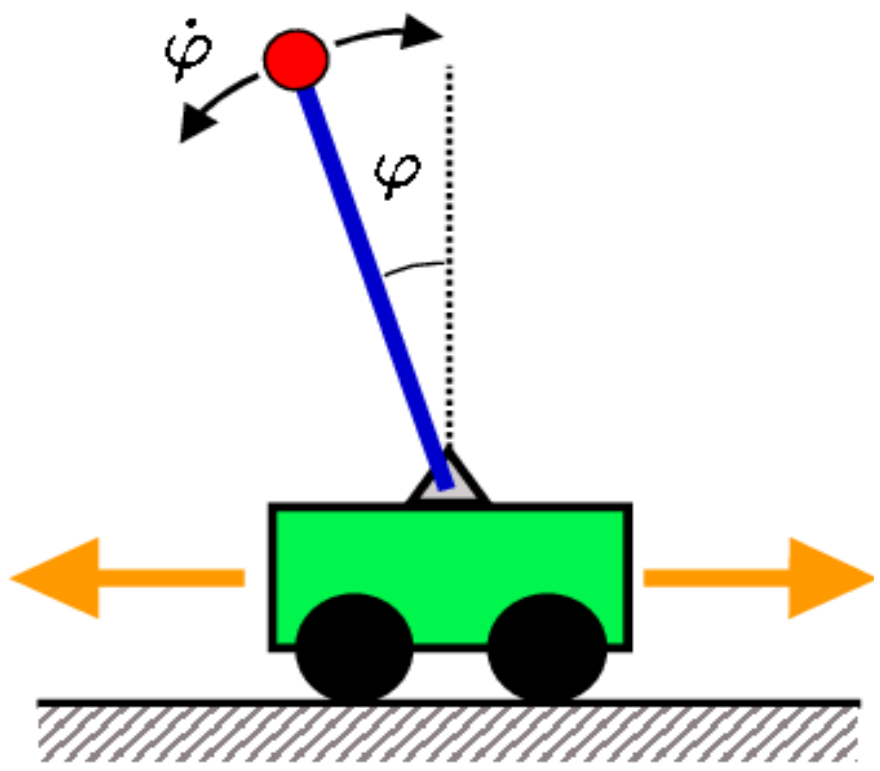

If we push the cart to the right, the pole will fall and the red ball will be at the lowest possible position, a single possible future state having a reward of zero. However, if we instead push it to the left, the pole will recover its up position, not only maximising the reward but also providing access to a greater number of future states.

The right decision here is then pushing to the left for two separate reasons:

1. It leads to a greater diversity of available future states.
2. It leads to future states with higher reward.

The process of scoring options needs then to be guided by a search for more possible future states, usually called 'exploration', while also taking into account how rewarding such future states are, i.e. 'exploitation'. Reaching and maintaining this fragile balance is the key idea behind any intelligent process.

We have nailed down the actual problem of intelligence to finding a way to scan the space of future states in such a way that exploration and exploitation are balanced during the process, and then make a decision about our next action based on the findings.

# 2 - Fundamental concepts

*"By far, the greatest danger of Artificial Intelligence is that people conclude too early that they understand it."*

**Eliezer Yudkowsky**

Building a detailed theory of intelligence based on these ideas requires fleshing out some fundamental concepts: the shape of this 'space of future states', what 'scanning' will mean to us, what 'balancing exploration and exploitation' means, and finally, how can we use information derived from the scanning process to make intelligent decisions over the available actions.

## 2.1 - Causal Cones

*"What is required is that the mind be prepared in advance, and be already stepping from thought to thought, so that it will not be too much held up when the path becomes slippery and treacherous."*

**Leibniz, on Rational Decision-Making**

In order to understand the 'space of future states' an intelligence will need to scan, we define a Causal Cone $X(x_0, \boldsymbol{\tau})$ as the set of all the paths the system can take starting from an initial state $x_0$ if allowed to evolve over a time interval of length $\boldsymbol{\tau}$, the 'time horizon' of the cone.

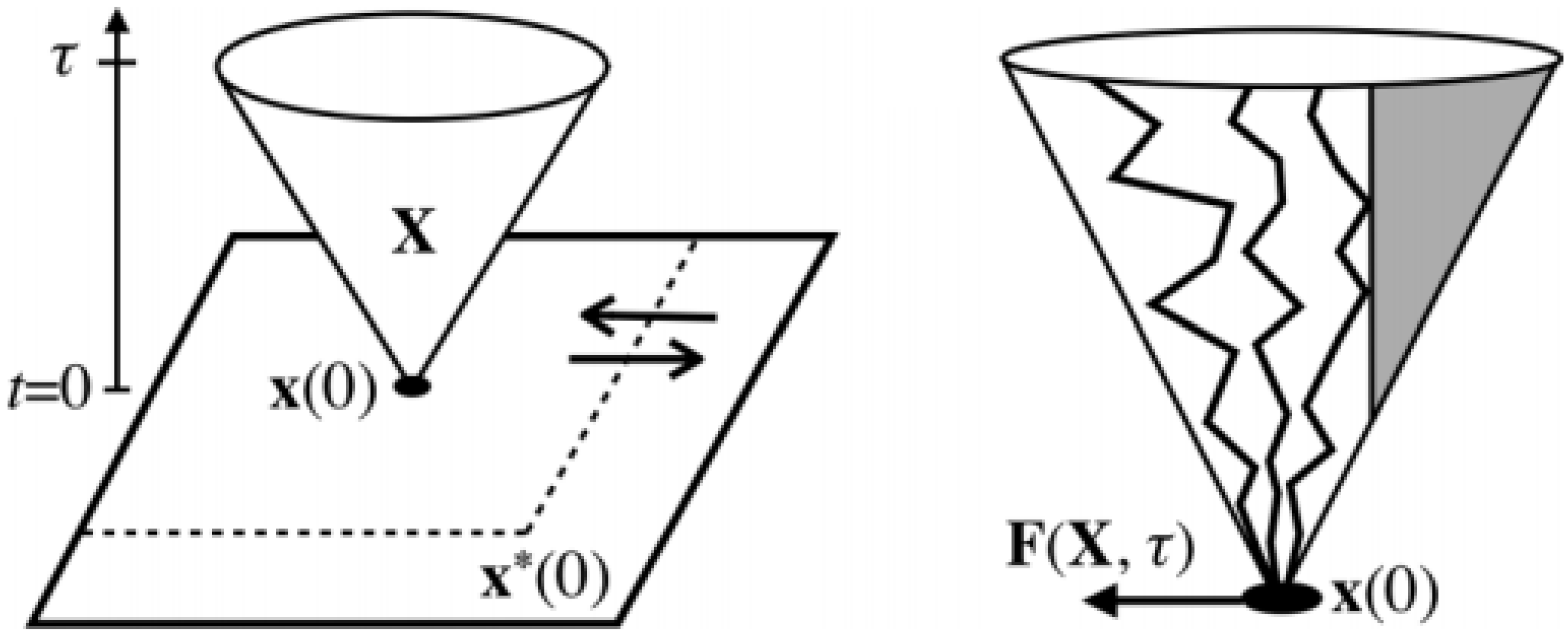

Causal cones are usually divided into two parts: the cone's 'horizon', formed by the final states ($t = \boldsymbol{\tau}$) for all the possible paths, and the rest of the cone ($t < \boldsymbol{\tau}$) usually referred to as the cone's 'bulk'.

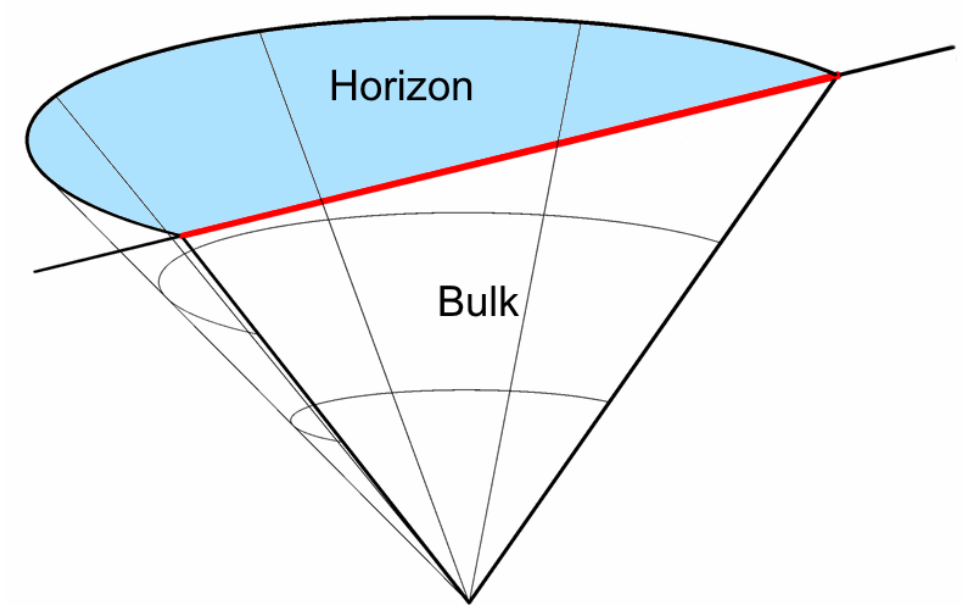

## 2.1.1 - Causal Slices

A Causal Slice of the cone $X(x_0, \tau)$ at time $t \in [0, \tau]$ is the horizon of the cone $X(x_0, t)$ which may be denoted by $X_H(x_0, t)$. Meanwhile, it is important to note that causal slices consist of a set of states, unlike causal cones that are formed by full paths.

We can imagine this causal slices as being the set of all future states the system can evolve to in a given time t, starting from $x_0$.

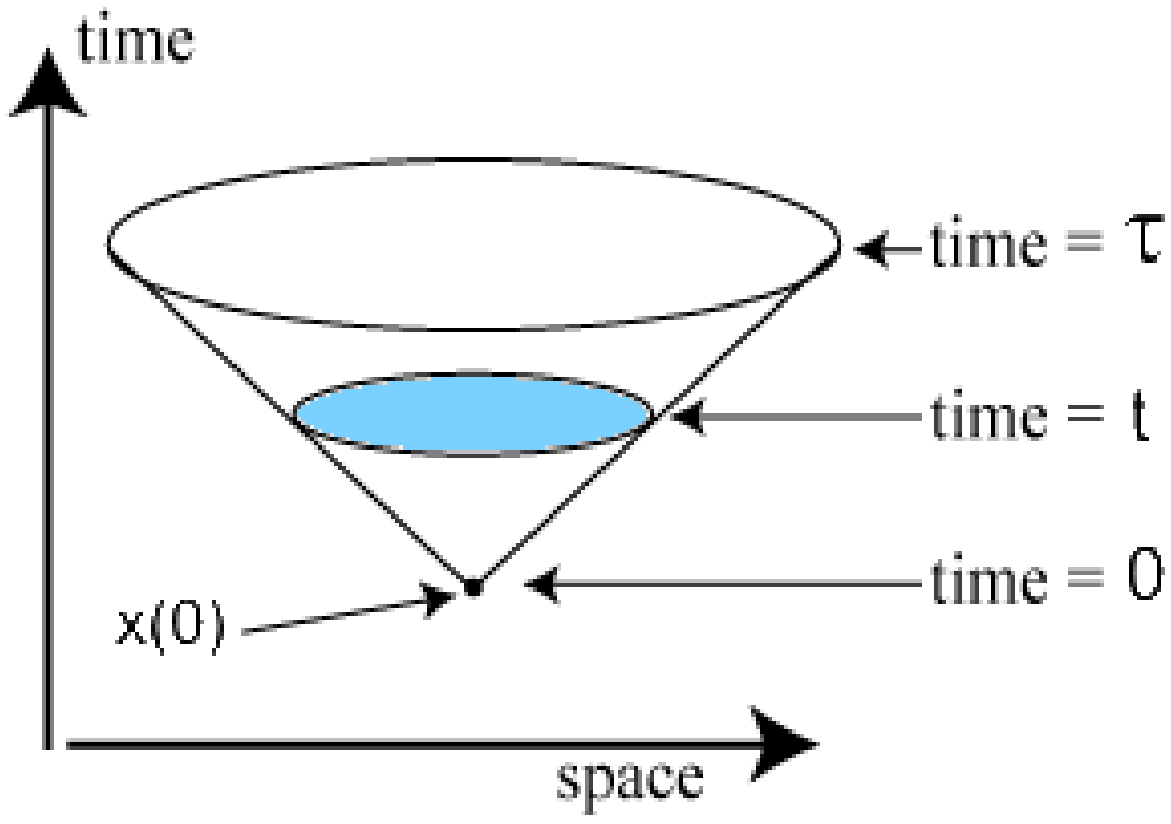

## 2.1.2 - Conditional Causal Cones

Given that the initial state $x_0$ contains specific information concerning the system's degrees of freedom, and given that the intelligence can alter those values by taking an initial action, all the paths forming the Causal Cone can then be partitioned based on this initial action taken.

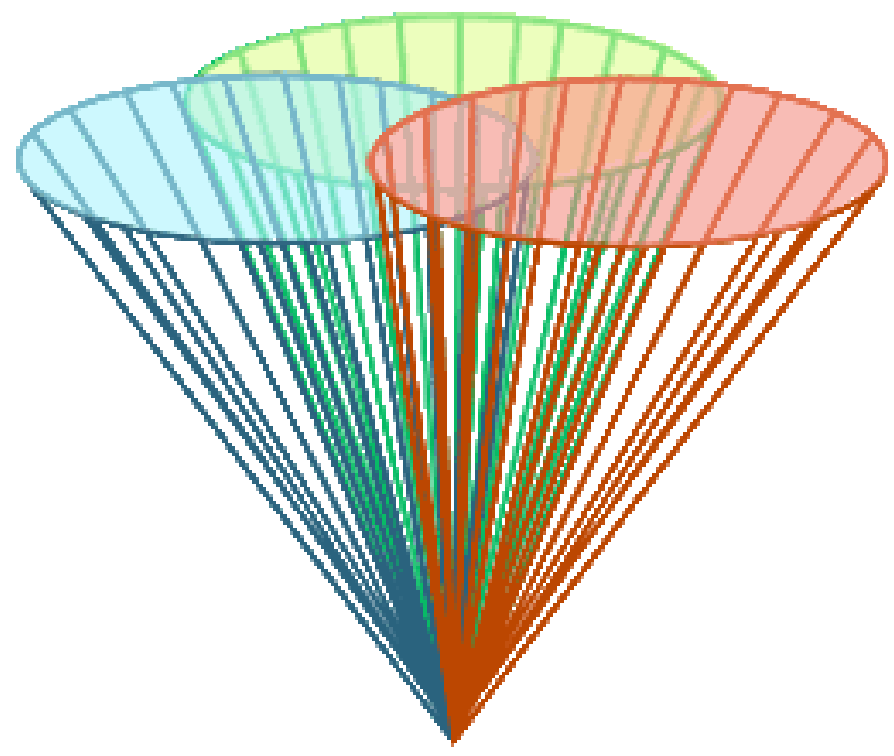

We define the Conditional Causal Cone associated with an action $a \in A$, $X(x_0, \boldsymbol{\tau}|a)$, as the cone formed by all the paths that start by taking the option a. The conditional cones are then a partition of the original cone:

$$X(x_0, \tau) = \bigcup_{a \in A} X(x_0, \tau|a)$$

## 2.2 - Reward function

> *"To employ the art of consequences, we need an art of bringing things to mind, another of estimating probabilities and, in addition, knowledge of how to evaluate goods and ills."*
>
> **Leibniz, on Rational Decision-Making**

In our setup, we will assume a reward function R(x) is defined over the state space. This reward function must follow some basic rules in order to be useful to the intelligence.

### 2.2.1 - Dead vs Alive states

We will say a state of the system is 'alive' from the intelligence standpoint when:

- It is a feasible state, meaning the system dynamics allow it.
- Modifying the degrees of freedom causes the system evolution to be affected.

We will then say the intelligence is 'alive' when it is in a alive state, and 'dead' in the other cases. This naturally defines the most basic form of reward function associated with the goal "keep alive":

$R_0(s) = 1$ if s is an 'alive' state
$R_0(s) = 0$ if s is an 'dead' state

In general, this basic reward function is always present, identifying the zones where you don't want your agent to be so the algorithm has to avoid visiting at any cost.

### 2.2.2 - Reward function properties

> *"I came to see that there is a species of mathematics in estimating reasons, where they sometimes have to be added, sometimes multiplied together in order to get the sum. This has not yet been noted by the logicians."*
>
> **Leibniz, on Rational Decision-Making**

A function R(x) defined over the state space of the system can be considered a reward function for an intelligent agent when it meets the following criteria:

1. The reward is positive for all alive states of the system.
2. The reward is zero for all dead states of the system.
3. States with higher rewards are considered better for the agent.

As an example, consider R(x) as being the battery level of an electric powered remote controlled agent, then:

1. When R(x) is positive, the batteries are working and the intelligence can take decisions that will affect the evolution of the system.
2. if R(x) is zero, the agent is out of batteries and any decision the intelligence could take will not affect the evolution of the system.
3. The higher the energy, the better for the agent.

Usually, our actual reward will be a composition of several more basic rewards: keep alive, energy level, health level, etc. that are multiplied together to build the reward being maximised by the agent.

$$R(s) = R_0(s) \times R_1(s) \times \dots R_{n-1}(s) \times R_n(s)$$

### 2.2.3 - Relativize: Universal reward reshaping

Sometimes a reward function is too badly shaped to be directly used: if you want the agent to get 'as much money as it can', you need to add a reward proportional to the agent's bank account balance, a figure that can easily be zero or even negative.

Whenever we have a reward component R(s) we cannot guarantee to strictly follow the above rules -as it would happen if reward could be negative for instance- we would need to reshape it in order to force the needed properties.

1. Get the mean and std. dev. of the rewards on the walkers states.

2. Normalize into N(0, 1) with $R_N(s) = (R(s) - Mean) / Std_dev$
3. Finally, define the new reward as:

$$R(s) = Exp(R_N(s)) \quad \text{if } R_N(s) \leq 0$$
$$R(s) = 1+Ln(1+R_N(s)) \quad \text{if } R_N(s) > 0$$

### 2.2.4 - Reward density over Causal Slices

For every slice $X_H(x_0, t)$ of the causal cone, we can calculate the total reward $R_{TOT}(x_0, t)$ of the slice as the integral of the reward over the slice. We may then convert the reward into a probability density $P_R$ over the slice as follows:

$$P_R(x|x_0, t) = R(x) / R_{TOT}(x_0, t)$$

## 2.3 - Policies: defining strategies

A policy $\pi$ is a set of two functions that completely define the strategy an intelligent system would follow when scanning the future consequences of its actions and deciding on the next action to take.

$$\pi = \{\pi_S, \pi_D\}$$

### 2.3.1 - Scanning policy

The scanning policy $\pi_S$ is a function that, given a state $x \in E$ of the system, outputs a probability distribution P over the available actions:

$$\pi_S : E \rightarrow P$$

When considering the probability of choosing a particular action a, we will use the conditional notation $\pi_S(a|x)$.

A special case of scanning policy is the random policy $\pi_S^{RND}$ that would assign a uniform distribution over the actions regardless of the state x, so all actions are equally probable and $\pi_S^{RND}(a_i|x) = \pi_S^{RND}(a_j|x)$, $\forall\, a_i, a_j \in A$:

$$\pi_S^{RND} : E \rightarrow P \text{ with P a uniform distribution.}$$

### 2.3.2 - Deciding policy

*"It is the mark of a truly intelligent person to be moved by statistics."*

**George Bernard Shaw**

A policy must also include a mechanism to score the available actions and assign them a probability of being chosen after the scanning phase is finished. This deciding probability distribution shall be denoted by:

$$\pi_D : A \rightarrow [0, 1]$$

There is also a special case where the decision is taken in a random manner so each action is equiprobable and $\pi_D^{RND}(a_i) = \pi_D^{RND}(a_j)$, $\forall\, a_i, a_j \in A$:

$$\pi_D^{RND} : A \rightarrow [0, 1] \text{ is a uniform distribution.}$$

## 2.3.3 - Probability density due to policy over Causal Slices

Once a scanning policy $\pi_S$ is defined, we may use it to calculate the probability of the system evolving to a given state, as the policy serves as a transition probability which, as in a Markov chain, allows us to project the evolution of the probability density over time.

For instance, in the initial causal slice for t = 0, the density is concentrated on the initial state $x_0$, as the slice $X_H(x_0, t = 0)$, only contain this state. However, as time increases, the causal slice volume will grow and the distribution will evolve.

We can then define this 'scanning probability density' $P_s$ as being a function defined over the slice $X_H(x_0, t)$, with parameters $x_0$, t and $\pi_S$:

$$P_s(x|x_0, t, \pi_S)$$

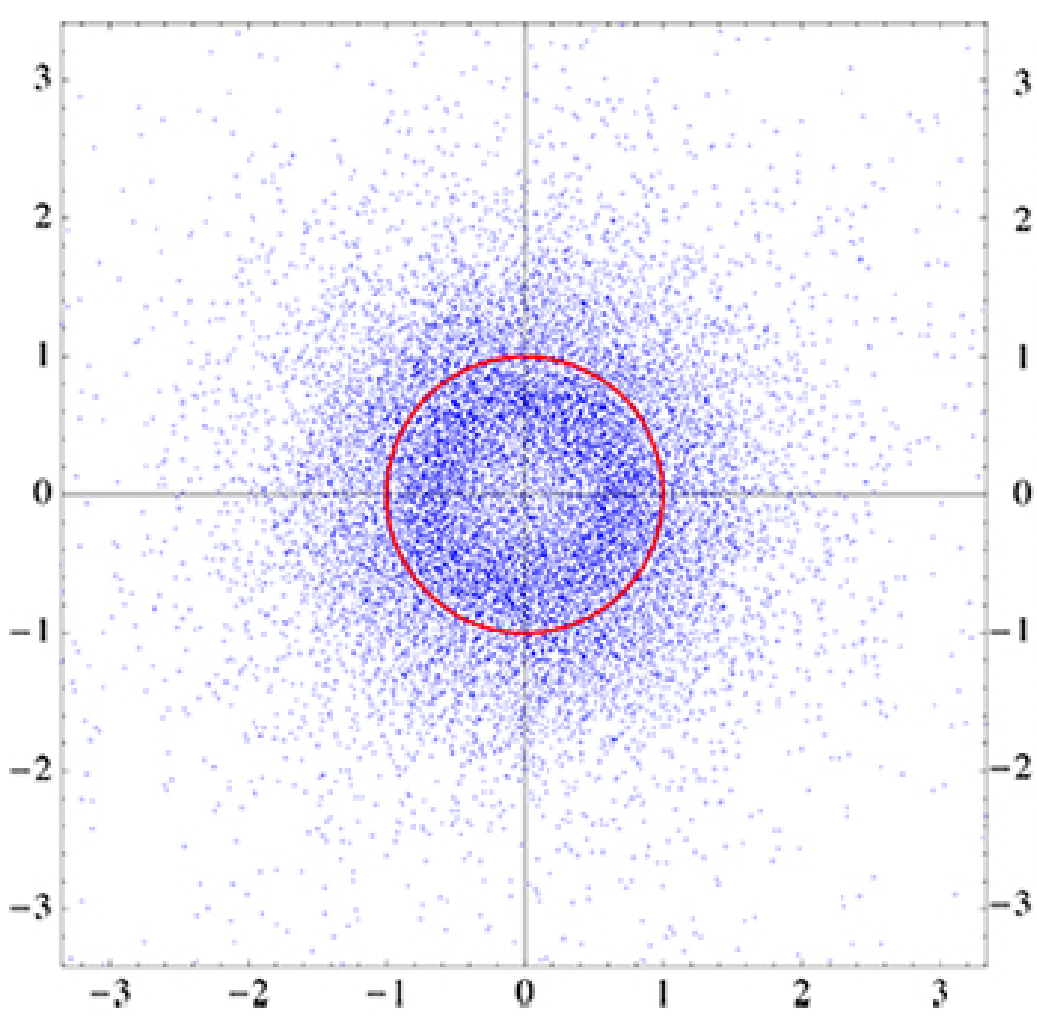

It is important to note that, given a scanning policy $\pi_S$, some of the states in a slice could be unreachable. For instance, if a policy chooses the same action all the time with probability 1, then $P_s(x|x_0, t, \pi_S)$ is not guaranteed to be non zero for all x in the slice, as it occurs with the random policy $\pi_S^{RND}$.

## 2.4 - Divergence between distributions

We will need to define a reliable measure of how similar two probability distributions are. This is usually done using the Kullback-Leibler divergence as follows:

$$D_{KL}(p \,||\, q) = -\boldsymbol{\Sigma}(p_i \,\mathrm{Log}(p_i/q_i)) \geq 0$$

This formulation requires $q_i > 0$ whenever $p_i > 0$. In our case, where $p = P_R(x|x_0, t)$ and $q_i = P_s(x|x_0, t, \pi_S)$, this is not guaranteed to be true as we noted before, but we could use Gibbs' theorem to find a divergence formulation which doesn't suffer from this weakness.

**Gibbs' Theorem 1.** Let P, Q $\in$ $\mathbf{P}_n = \{p \in \mathbb{R}^n : p_i \geq 0, \ \boldsymbol{\Sigma}\, p_i = 1\}$, then $\boldsymbol{\Pi}(q_i^{pi})$ is maximized by $\boldsymbol{\Pi}(p_i^{pi})$.

Using this theorem we may define a different divergence of two distributions as follows:

$$D_H(P \,||\, Q) = \mathrm{Log}(\boldsymbol{\Pi}(p_i^{pi}) \,/\, \boldsymbol{\Pi}(q_i^{pi}))$$

This divergence is well defined for any possible distributions p and q, including the problematic case when ($p_i > 0$, $q_i = 0$) and it satisfies the main properties of the KL-divergence we need:

1. $D_H(P \,||\, Q)$ is well defined for any pair of distributions.
2. $D_H(P \,||\, Q) \geq 0$ for any pair of distributions.
3. $D_H(P \,||\, Q) = 0$ if and only if $p = q$.

# 3 - Defining Intelligence

*"There is always another way to say the same thing that doesn't look at all like the way you said it before. I don't know what the reason for this is. I think it is somehow a representation of the simplicity of nature."*

**Richard P. Feynman**

We will define intelligence as the ability to minimize a 'sub-optimality' coefficient based on the similarity of two pairs of probability distributions obtained during the processes involved: scanning and the decision-making.

## 3.1 - Scanning process

*"Intelligence is not to make no mistakes, but quickly to see how to make them good."*

**Bertolt Brecht**

In the scanning phase, the possible future outcomes of the initial actions are sampled using the scanning policy $\pi_S$.

### 3.1.1 - Causal Slice divergence

Given a slice of the causal cone $X_H(x_0, t)$ and a scanning policy $\pi_S$ we previously defined two probability distributions over it: the scanning density distribution $P_s(x|x_0, t, \pi_S)$, and the reward distribution $P_R(x|x_0, t)$.

The reward distribution is provided by the environment in an objective manner so the policy can not change it, while the scanning density distribution is directly dependent on $\pi_S$, so it makes sense to use the divergence $D_H(P_R, P_s)$ as a measure of how well the scanning policy leads the agent to rewarding states.

### 3.1.2 - Intelligent Scanning

*"...in order to decide what we ought to do to obtain some good or avoid some harm, it is necessary to consider not only the good or harm in itself, but also the probability that it will or will*

*not occur, and to view geometrically the proportion all these things have when taken together.”*

**Leibniz, on Rational Decision-Making**

We will define the ‘Intelligent Scanning’ as the optimal policy $\pi_S^{OPT}$ that produces a scanning density distribution that is proportional to the reward for every slice of the causal cone:

$$P_s(x|x_0, t, \pi_S) \propto R(x),\ \forall x \in X_H(x_0, t)$$

This implies that both probability densities are coincident:

$$P_s(x|x_0, t, \pi_S) = P_R(x|x_0, t),\ \forall t \in [0, \boldsymbol{\tau}], \forall x \ \in X_H(x_0, t)$$

Equivalently, we may define $\pi_S^{OPT}$ as the policy that makes the causal slice divergence to equal zero for any t $\in$ [0, $\boldsymbol{\tau}$]:

$$D_H(P_R(x \mid x_0, t), P_s(x \mid x_0, t, \pi_S)) = 0,\ \forall t \in [0, \boldsymbol{\tau}], \forall x \in X_H(x_0, t)$$

The idea behind this definition is that the optimal way of scanning a space is to make the probability of searching on a particular zone to be proportional to the expected reward: should you be searching for gold over a wide landscape, it would make sense to adjust the density of gold-miners in different zones to be proportional to the density -probability of finding- gold.

### 3.1.3 - Scanning sub-optimality coefficient

The more similar the two distributions are, the more intelligent and efficient the scanning will be, and, in the limit when the policy is optimal, we would reach an equilibrium where the divergence between both distributions is exactly zero.

Given that real-world scanning policies will not produce scanning probability densities exactly proportional to the rewards, this divergence will generally not be exactly zero over the different slices $X_H(x_0, t)$, so integrating the divergences over time can provide us with a measure of the sub-optimality of the scanning policy:

$$Scan(\pi_s|x_0, \tau) = \int_{t=0}^{\tau} D_H(P_R(x|x_0,\ t), P_s(x|x_0,\ t,\ \pi_s))\ dt$$

In order to scale this coefficient into a more sensible figure, we may define the ‘unit of sub-optimality’ to be the sub-optimality associated with the random scanning policy $\pi_S^{RND}$.

$$\text{Random Scan Sub-optimality} = Scan(\pi_S^{RND}|x_0, \boldsymbol{\tau})$$

We may now divide the previous coefficient into this sub-optimality unit, so only policies that are ‘better than random’ will score below 1, while ‘worse than random’ ones will score above 1.

$$\text{Scan Sub-Optimality}(\pi_S|x_0, \boldsymbol{\tau}) = \text{Scan}(\pi_S|x_0, \boldsymbol{\tau}) \;/\; \text{Scan}(\pi_S^{RND}|x_0, \boldsymbol{\tau})$$

# 3.2 - Decision process

*“There is hardly anyone who could work out the entire table of pros and cons in any deliberation, that is, who could not only enumerate the expedient and inexpedient aspects but also weigh them rightly. Now, however, our characteristic will reduce the whole to numbers, so that reasons can also be weighed, as if by a kind of statics.”*

**Leibniz, on Rational Decision-Making**

As the second part of the process, the information obtained by the scanning phase is used to decide which action to take or, more generally, the probability $\pi_D(a)$ of each action to be taken.

## 3.2.1 - Intelligent decision

We will say a decision, defined as a probability distribution $\pi_D(a)$ over all possible actions $a \in A$, is the ‘intelligent decision’ $ID(a|\pi_S, \boldsymbol{\tau})$ for policy $\pi_S$ and time horizon $\boldsymbol{\tau}$, when the probabilities are proportional to the entropy of the conditional scanning probability densities for the actions over the last slice of the cone, $X_H(x_0, \boldsymbol{\tau})$.

$$ID(a|\pi_S, \boldsymbol{\tau}) \propto \mathcal{H}(P_s(x|x_0, \boldsymbol{\tau}, \pi_S, a))$$

Using that the conditional probabilities $P_S$ for different actions $a \in A$ are independent and that the corresponding conditional causal cones form a partition of the causal cone, we have that:

$$\mathcal{H}(P_s(x|x_0, \boldsymbol{\tau}, \pi_S)) = \sum_{a \in A} \mathcal{H}(P_s(x|x_0, \boldsymbol{\tau}, \pi_S, a))$$

So we may obtain the intelligent decision as a probability density over the actions as follows:

$$ID(a|\pi_S, \boldsymbol{\tau}) = \mathcal{H}(P_s(x|x_0, \boldsymbol{\tau}, \pi_S, a)) \;/\; \mathcal{H}(P_s(x|x_0, \boldsymbol{\tau}, \pi_S))$$

This formulation implies that the intelligent decision in the discrete and continuous cases are:

| Discrete case | Intelligent decision= arg max $ID(a|\pi_S, \boldsymbol{\tau})$ |
|---|---|

| Continuous case | Intelligent decision= $\int_{a \in A} a \cdot ID(a \vert \pi_s, \tau)\, da$ |
|---|---|

### 3.2.2 - Decision sub-optimality coefficient

Given a policy $\pi = \{\pi_S, \pi_D\}$ that generates a probability distribution $\pi_D(a)$ over the actions, we can define its sub-optimality as the divergence with the ideal distribution ID(a):

$$\text{Decision sub-optimality}(\pi|x_0, \boldsymbol{\tau}) \propto D_H(ID(a|\pi_S, \boldsymbol{\tau}), \pi_D(a))$$

As we want this coefficient to be 1 for the random policy, we can define our unit of sub-optimality as the sub-optimality of the random decision policy $\pi_D^{RND}$ , the uniform distribution:

$$\text{Decision sub-optimality}(\pi|x_0, \boldsymbol{\tau}) = D_H(ID(a|\pi_S, \boldsymbol{\tau}), \pi_D) / D_H(ID(a|\pi_S, \boldsymbol{\tau}), \pi_D^{RND})$$

## 3.3 - Global sub-optimality

Given a policy $\pi$ responsible for both scanning and deciding, we can define its global sub-optimality as the average of both sub-optimality coefficients:

$$\text{Sub-optimality}(\pi|x_0, \boldsymbol{\tau}) = (\text{Scan sub-optimality}(\pi_S|x_0, \boldsymbol{\tau}) + \text{Decision sub-optimality}(\pi|x_0, \boldsymbol{\tau})) / 2$$

This global sub-optimality coefficient allows us to determine which policies are approximately random (≈1) and which are nearly optimal (≈0).

## 3.4 - Policy IQ

Given the definition of sub-optimality of a policy, we can define the IQ of a given policy as follows:

$$IQ(\pi|x_0, \boldsymbol{\tau}) = 1 / \text{sub-optimality}(\pi|x_0, \boldsymbol{\tau})$$

By using a moving average of this IQ over time as the system evolves, we can build a reliable real-time measurement of the intelligence of a particular system evolution.

# 4 - FMC: a Fractal Monte Carlo algorithm

> *"Intelligence is the ability to avoid doing work, yet getting the work done."*
>
> **Linus Torvalds**

Once we have a sensible definition of intelligence, the next step is to use it to define a practical and efficient "decision-taking" -also known as "planning"- algorithm, or, more precisely:

"Given (1) a system with some degrees of freedom that can be controlled, (2) an informative simulation of the system's dynamics (not necessarily a perfect simulation nor a deterministic one), and (3) a reward function defined over the state space, find an algorithm that use this information to push the degrees of freedom in such a way that the system behaves -or evolves- intelligently".

We will guide our search in two ambitious "design principles":

1. An algorithm with a sub-optimality coefficient tending to zero.
2. An algorithm with the lowest time-complexity possible.

The algorithm presented here is not the only possible one, nor it is a complete implementation of all the potential uses of the previous theory. It must instead be considered as one direct and intuitive use of the concepts presented in order to generate intelligent behaviour from the raw simulation of a system.

## 4.1 - Algorithm blueprints

*"You can recognize truth by its beauty and simplicity. When you get it right, it is obvious that it is right—at least if you have any experience—because usually what happens is that more comes out than goes in. (...) the truth always turns out to be simpler than you thought."*

**Richard P. Feynman**

Before presenting the pseudo-code of the algorithm, we will introduce the ideas behind it and how its design aims to meet the previously defined theory in the lowest computational time complexity.

### 4.1.1 - Starting at Monte Carlo

*"Any one who considers arithmetical methods of producing random digits is, of course, in a state of sin."*

**John von Neumann**

We will start our implementation by considering the random policy or, equivalently, using the standard Monte Carlo approach, where a set of independent random walks are simulated over a number of discrete time steps to actually build a collection of paths in the causal cone considered in the theory.

In practical terms, we will need an informative simulation function -not necessary perfect nor deterministic- that, given a system state -that includes the positions of the different degrees of freedom the AI can modify- and a small delta of time dt, outputs the expected next state of the system:

$$x(t+dt) = Simulation(x(t), dt)$$

Starting at the actual system's state $x_0$ and iterating the process of taking a random decision over the degrees of freedom and then simulating the next state for a fixed number T of ticks , we can build a random walk of length $\tau$ = T*dt.

In the image below, two different actions -pushing buttons "A" or "B"- are available. In the continuous case, actions would be real vectors representing the velocities of the degrees of freedom, or how strong it pushes each free param, and in which direction.

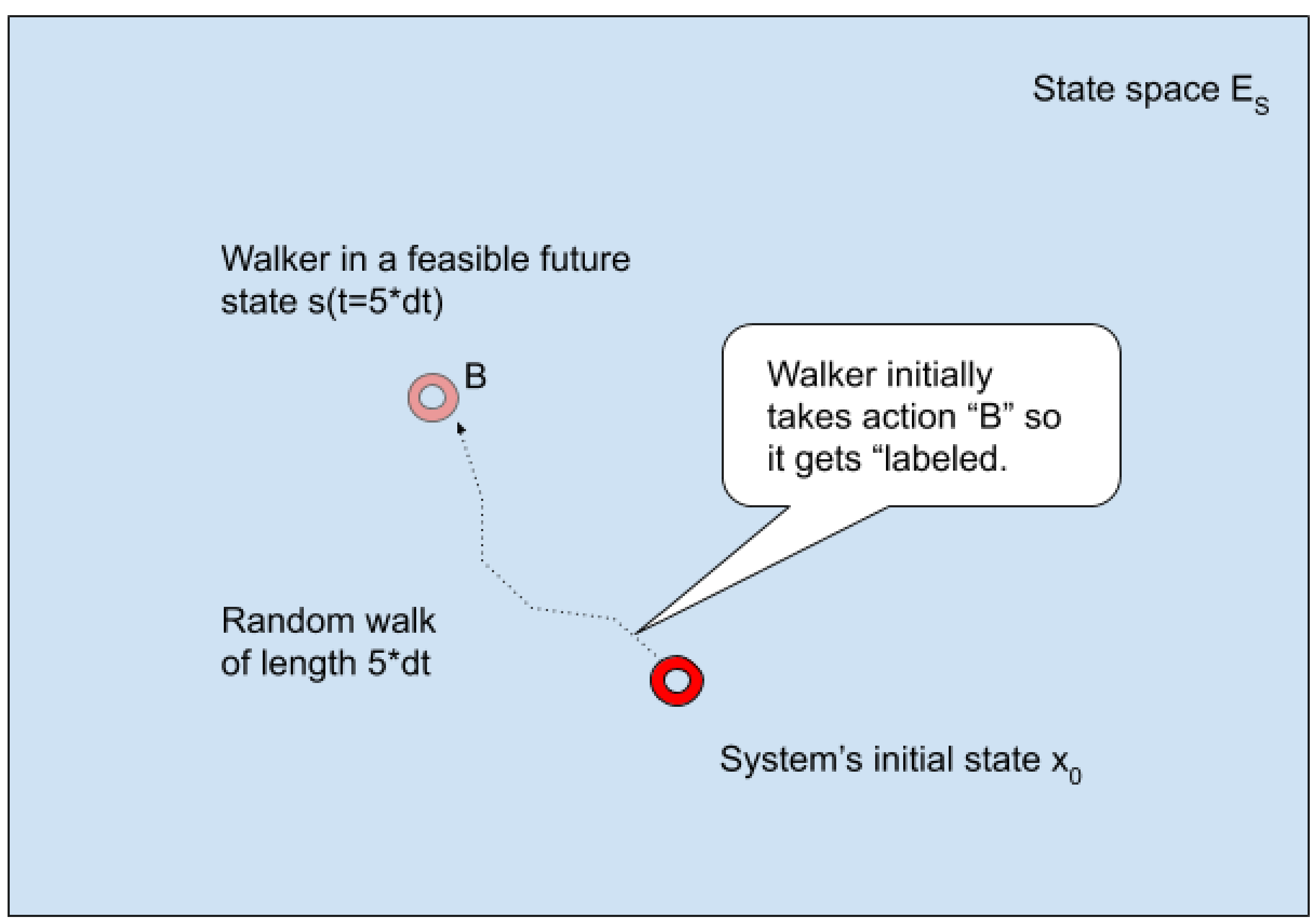

As we are looking for a decision-taking algorithm, the first decision -or action- taken during the walk, "B" in this example, will be an important piece of information, so we will keep it in the walker's internal state for later use.

By building one path after another we can actually build a set of N 'feasible random walks' starting at $x_0$ and ending at one of the possible system future states.

## 4.1.2 - Choosing the intelligence parameters

At this point, some practical questions like how many random walks we will be using, the time horizon those walks will explore, and the number of ticks we will divide this time into will probably arise.

Before going any deeper in the algorithm it is worth spending some lines addressing those questions as they are the main parameters related to the CPU you will need and the quality of the results.

### 4.1.2.1 - Decisions per second

The first parameter to be chosen is about the number of decisions per second the agent will be taking, or the FPS (frames per second) of the algorithm. We are basically deciding on the length of the dt used in the simulations, our tick length.

Basically, this length depends on the system reaction times. If our agent is modeling a fly we will need a faster decision taking, so a lower tick length. If it were modeling a spaceship traveling to andromeda, we could safely take one decision per month or even year.

As a guiding number, human brains are considered to run at about 12 decisions per second, so if we were to mimic some human behaviour, a dt of about 0.1 seconds (or 10 FPS) is a nice starting point.

Setting a FPS higher that the reaction time will not really improve the results, while CPU time will grow proportionally to FPS, so just keep this figure around the sweet point.

#### 4.1.2.2 - Time horizon

The time horizon dictates how far into the future will the walkers explore the consequences of their initial actions.

The idea here is to try to scan long enough to detect the problems before you can not avoid them and, again, it depends naturally on the task:

- A F1 driver, deciding on the driving wheel and pedals, needs to foresee where the car will be in about 5-10 seconds in order to properly drive a race.
- A spaceship traveling to andromeda needs to foresee some years to know if pushing the thruster now will lead you to andromeda or not.

#### 4.1.2.3 - Number of walkers

This one is easy: the more the better. Of course it will use CPU linearly, but the most efficient way to improve the agent behaviour is using more walkers.

### 4.1.3 - Simultaneous walks

As the theory mandates that the 'density of scanning' -or 'density of walkers'- must be kept proportional to some reward density, we will need to build all the random walks simultaneously, so the density of walkers can be defined.

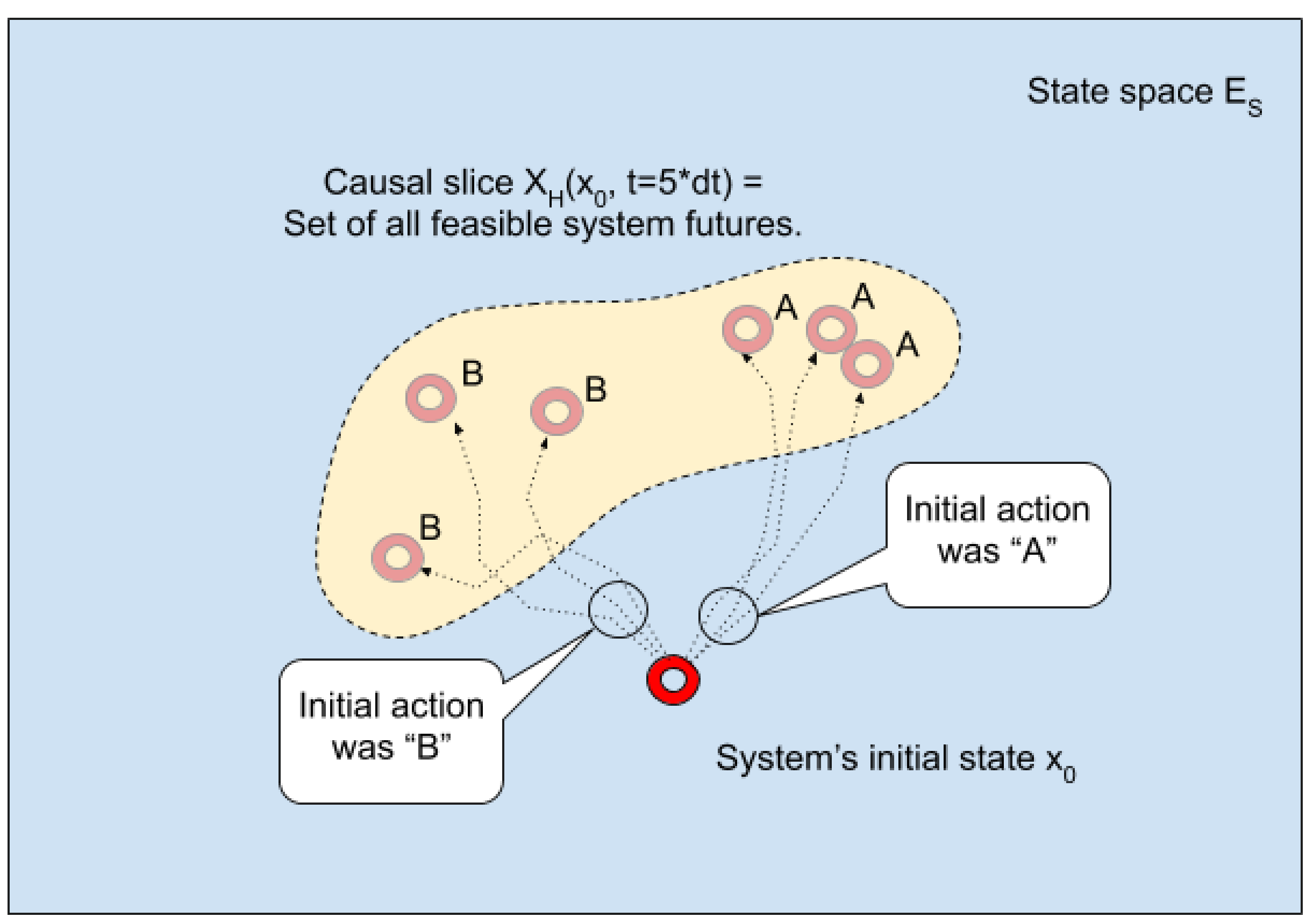

Before going any further, we will show a simple pseudo-code that would generate this set of walks and finally decide based on the highest reward found at the final states of each path: if the path leading to the most rewarding state found started with the action "A", then it will actually take action "A".

---

```
// INITIALIZATION:
// Create N walkers with copies of the system's state:
FOR i:= 1 TO N DO BEGIN
      // Walkers start at the system's initial state:
      Walker(i).State:= System.State
      // Take walker's initial decision:
      Walker(i).Initial_decision:= random values
END
// SCANNING PHASE:
// Evolve walkers from time=t to t+Tau in M ticks:
FOR t:= 1 TO M DO BEGIN
      // PERTURBATION:
      FOR i:= 1 TO N DO BEGIN
            // First tick use the stored initial decision
            IF (t=1) THEN
                  Walker(i).Degrees_of_freedom:= Walker(i).Initial_decision
            ELSE
                  Walker(i).Degrees_of_freedom:= random values
            // Use the simulation to fill the other state's component:
            Walker(i).State:= Simulation(Walker(i).State, dt:= Tau/M)
```

```
        END
END
// DECIDING PHASE:
Best:= ArgMax(Reward(Walker(i).State))
Decision:= Walker(Best).Initial_decision
```

---

Being this code just a simple starting point, it already meets some of our design goals:

1. The more accurate the simulation is, and the smaller that dt is made, the more compatible with the system's dynamic the generated paths are.
2. The bigger the number of walkers used, the bigger portion of the causal cone will be scanned.

## 4.1.4 - Probability densities

---

*"In applying dynamical principles to the motion of immense numbers of atoms, the limitation of our faculties forces us to abandon the attempt to express the exact history of each atom, and to be content with estimating the average condition of a group of atoms large enough to be visible. This method... which I may call the statistical method, and which in the present state of our knowledge is the only available method of studying the properties of real bodies, involves an abandonment of strict dynamical principles, and an adoption of the mathematical methods belonging to the theory of probability."*

---

**James Clerk Maxwell**

According to theory, no matter which partition $\{A_1, \dots , A_N\}$ of the causal slice you consider, the walker density $D_i$ on each part $A_i$ should be made proportional to the density of reward $R_i$

Our next step will be properly defining both densities.

### 4.1.4.1 - Density of walkers

In the next image we have partitioned the space with boxes of the same size. Each one of the four populated boxes $A_1$ to $A_4$ have a number of walkers $W_i$ inside it from a total of six walkers considered, so the walker's density at $A_i$ will be $D_i = W_i/6$ .

Please note that we didn't need it to be a partition, the different zones may overlap, so we are actually using a covering of the set of all the walker's positions.

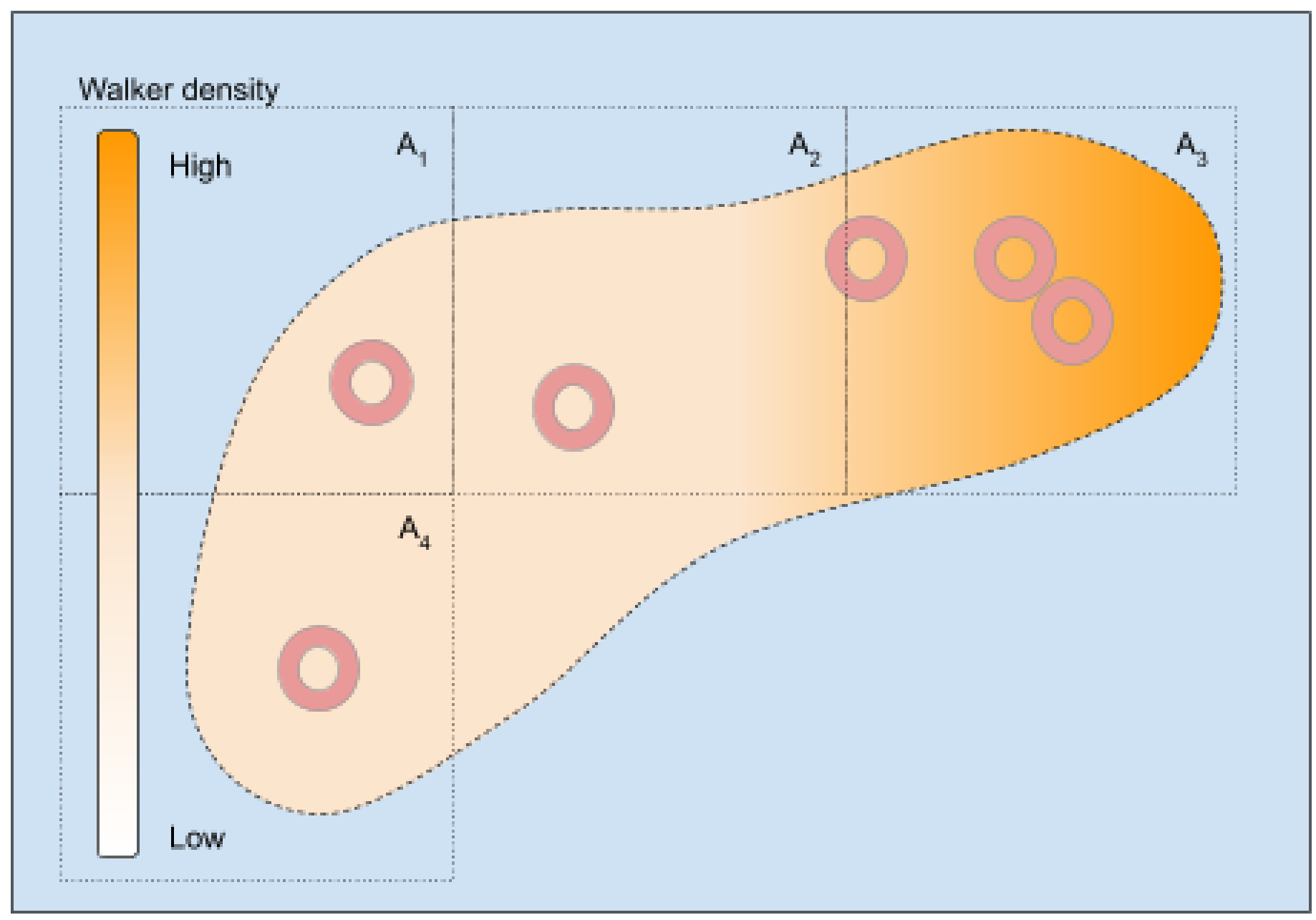

### 4.1.4.2 - Reward density

At the same time, a reward value is defined over the state space, so we can assign a reward value to each box $A_i$ by averaging the rewards at the positions of the walkers in $A_i$.

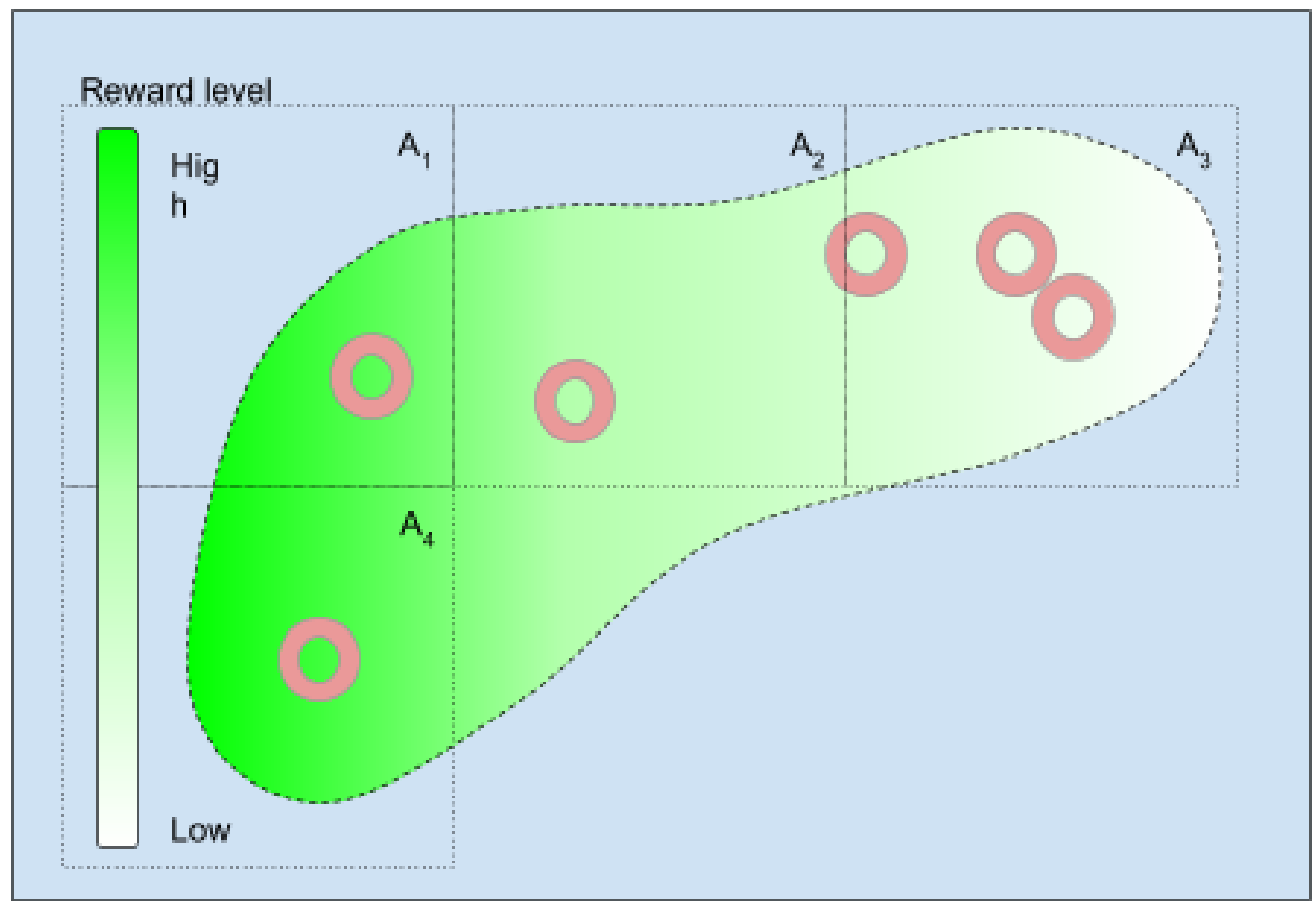

## 4.1.5 - Migratory flows

Our final goal is to make both densities proportionals, or, equivalently, we need $R_i/D_i$ to be as constant as possible.

Basically, $R_i/D_i$ is a measure of the "reward per capita" that a walkers in the box Ai will receive, and the idea of making this a constant could be interpreted as the need of a fair wealth distribution over the walkers: if zone $A_i$ has a much higher reward per capita than zone $A_j$ then walkers in $A_j$ will likely prefer to move to $A_i$. As we aim to make both densities to be proportional, we need to define a 'migratory flow' in order to balance the reward per capita.

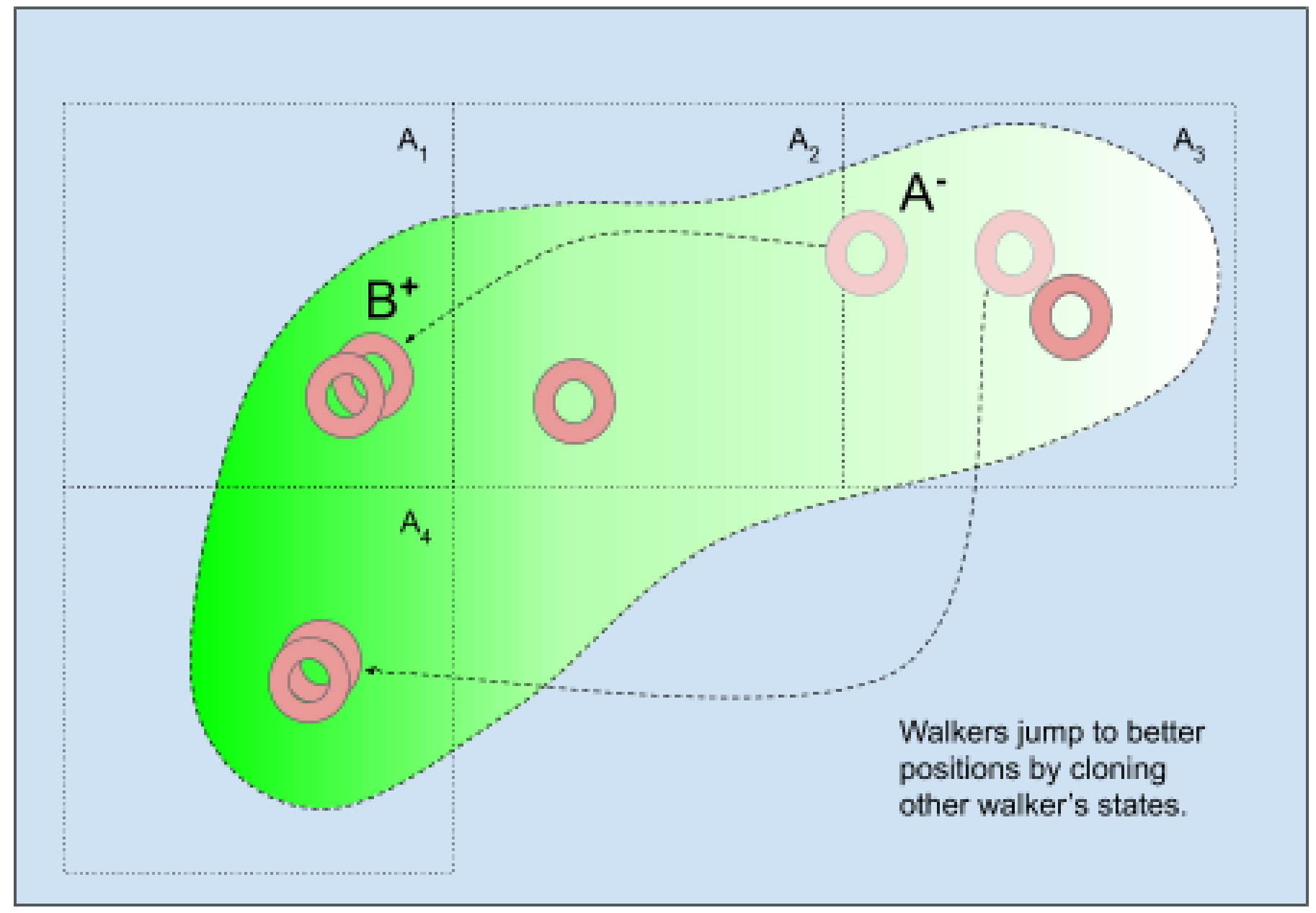

This migratory flow will need to detect zones where walker density is higher-than-proportional to the reward (zones with a low 'reward per capita') and move some walkers from there to other zones with higher 'reward per capita'.

In order to move a walker $W_1$ from zone $A_i$ to $A_j$ we could just select a random walker $W_2$ at $A_j$ and copy its state, $W_1$.state ← $W_2$.state. We say that $W_1$ 'cloned' into $W_2$ position.

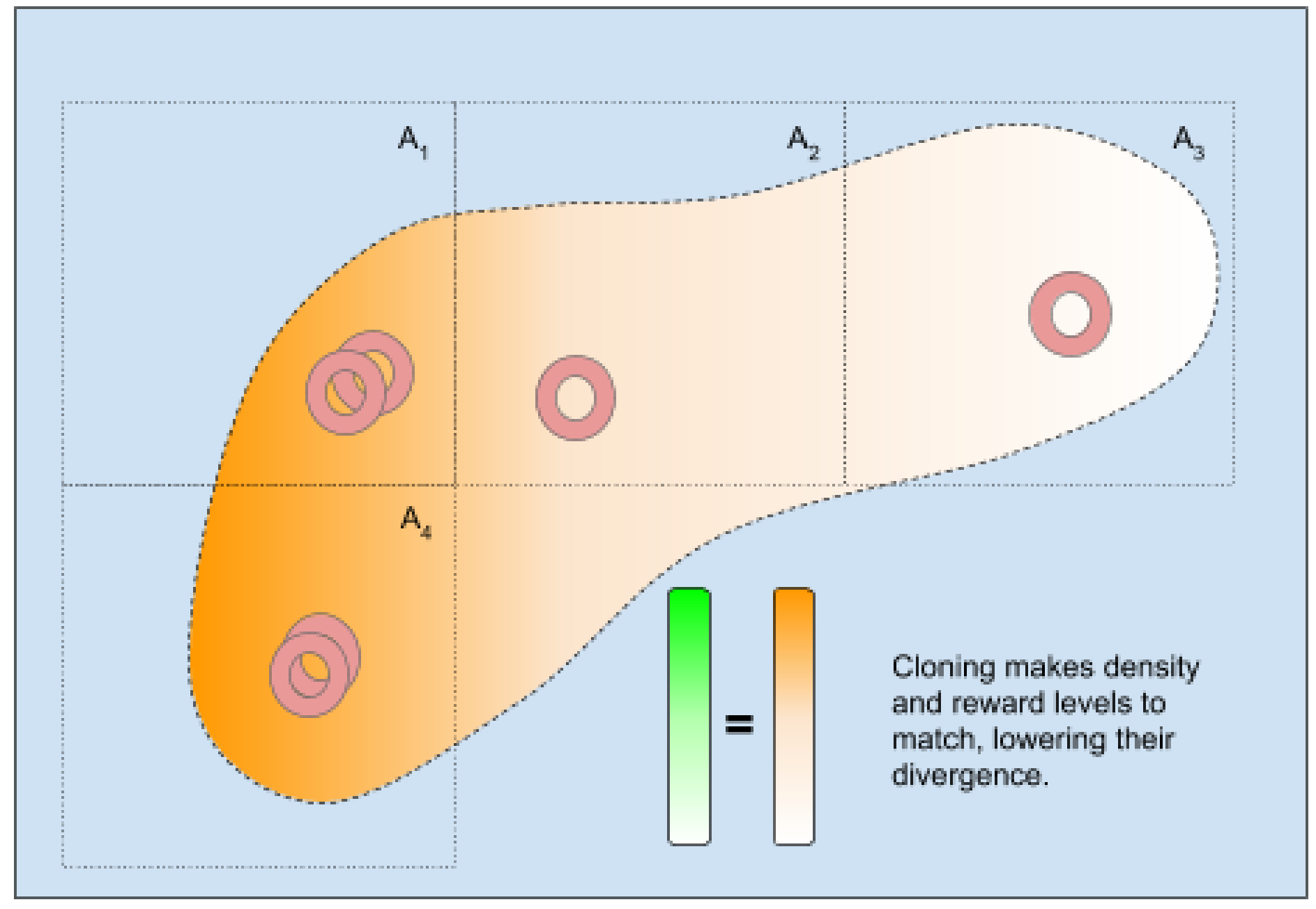

Here we just sketched a nïve way to choose which walkers needs to be cloned into which others by using boxes. In the final algorithm this idea will be replaced with a more general solution.

## 4.1.6 - Making the decision

*"Deliberation is nothing else but a weighing, as it were on scales, the conveniences and inconveniences of the fact we are attempting."*

**Leibniz, on Rational Decision-Making**

In the previous example, when a walker labeled with the first option "A" is cloned to the position of a walker labeled "B", it not only clone its position but also its label, so after the clone, the initial action "A" will have one 'follower' less, while action "B" will gain one. After some ticks are performed, the distribution of the initial actions in the population of walkers will vary.

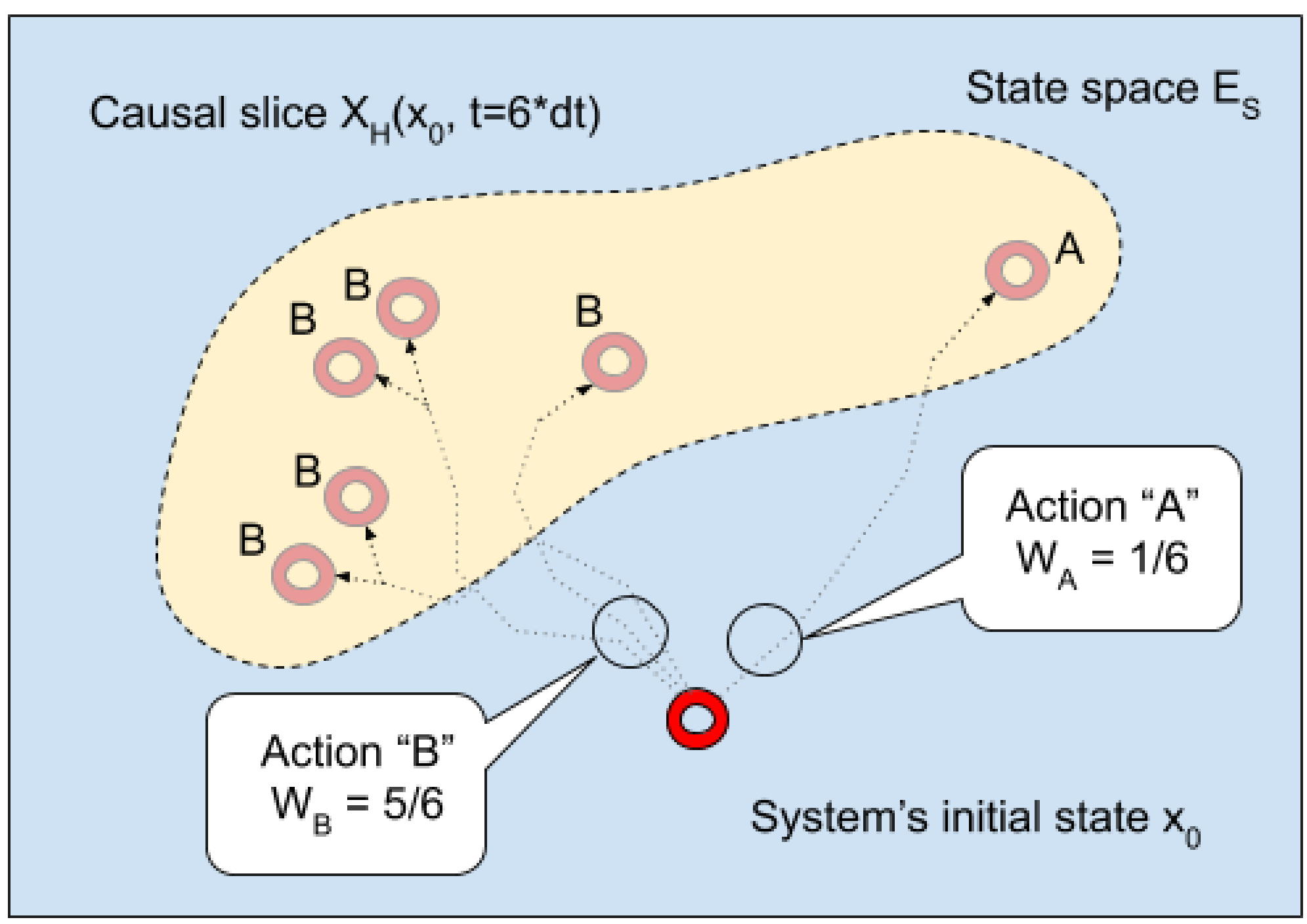

In the example above, both actions start having a proportion of 3/6 = 0.5 of the walkers, but after the migratory flow takes place, "B" population grows to 5/6 while "A" drops to 1/6.

If the process was to be stop here -time horizon $\tau$ was 6 ticks of length dt- then we would use these proportions to build our decision.

In the discrete case, the AI would take its decision by sampling an action from the probability distribution (1/6, 5/6) of the actions, so the most probable action would be "B".

In the continuous case, where actions are real vectors and there is no finite list of available actions, the decision is formed by averaging the initial decisions of all the walkers.

## 4.2 - The migratory process

We commented on the need of forcing migratory flows from areas were reward was low or population density high, but defining a density is a tricky thing that can also be computationally demanding.

The idea of using boxes -as in the previous introduction- was just a sketch of the idea, we didn't define a method to effectively choose which walkers will clone and, more importantly, which walkers will they copy from.

### 4.2.1 - Virtual reward

*"Presuming that a man has wisdom of the third degree and power in the fourth, his total estimation would be twelve and not seven, since wisdom be of assistance to power."*

**Leibniz, on Rational Decision-Making**

The 'Virtual reward' is a generalization of the 'reward per capita' concept introduced before, but instead of using an externally defined partition, we will build one with a different zone per walker, so we can obtain a 'personalized' reward value that will tell a walker how 'lucky' he is as compared to other walkers: the higher the reward is, and the fewer the walkers around you, the better.

Choosing one partition per walker to account for the number of walkers inside each part is actually just a way to approximate the concept of 'density of walkers' around a position. At this point we will just assume we have defined some general measure of such density around the position of each walker:

$D_i$ =Density of Walkers around $W_i$ position/state.

We will then define the 'virtual reward' $VR_i$ at a walker's $W_i$ position, where the reward value is $R_i$, as being:

$$VR_i = R_i / D_i$$

One simple way to define a density of walkers around a given position is to consider our covering as being formed by a set of spheres of a fixed radius in the state space, each centered at one walker position, and counting the number of walkers inside each one to get a density.

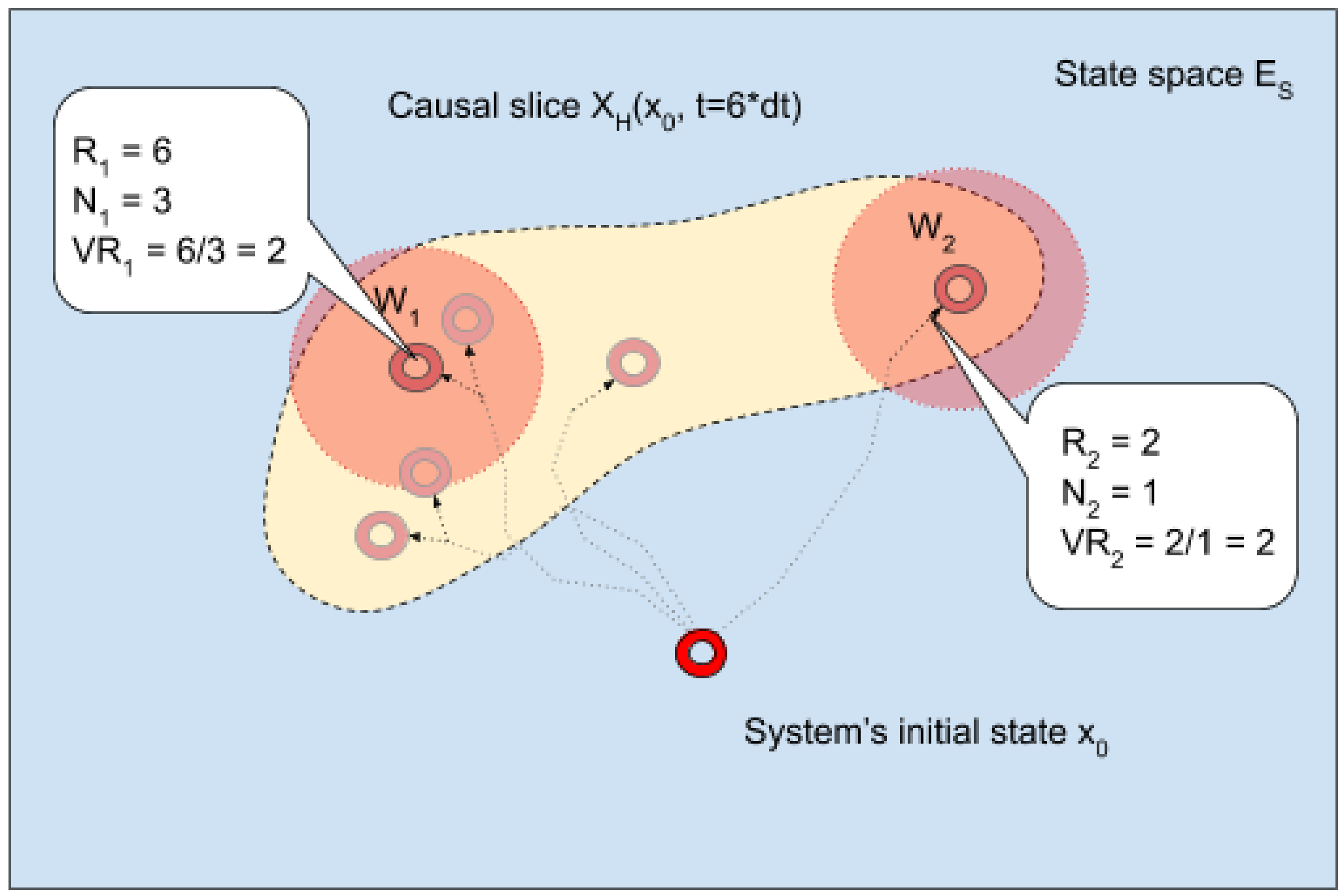

Please note that, as we will only focus on the proportionality of those densities -and not in their actual values- we can safely use the number $N_i$ of walkers inside the ball as a density, without dividing it by N, as it will not alter the proportions.

Then, by looking at the image, we can say that at walker $W_1$ position, there is a reward of $R_1 = 6$ and a walker density of $N_1 = 3$, so we will calculate its virtual reward as:

$$VR_1 = R_1 / N_1 = 6 / 3 = 2$$

If we compare it with the virtual reward for walker $W_2$, $VR_2 = R_2 / N_2 = 2 / 1 = 2$, we find that both positions are equally 'appealing' as the number of walkers is kept proportional to the reward.

## 4.2.2 - Simplifying the Virtual Reward

*"Everything should be made as simple as possible,*
*but not simpler."*

**Albert Einstein**

Using densities around the walker positions may not be an ideal approach for several reasons:

1. The algorithm would need to check the distances between all possible pairs of walkers, making the computational time-complexity to be, at least, of order $O(n^2)$, where n accounts for the number of walkers times the number of ticks you divide your time horizon in.

2. We would need to externally set a radius that makes sense and, probably, adjust it dynamically so it doesn't get too big or small during the process. You could, for instance, force the average number of walkers per ball to be between some reasonable values, let's say in (1.5, 3), so if radius is so small that there is only one walker at each ball then, as 1 < 1.5, you would need to increase the radius.

Our first simplification will eliminate the need for an externally defined radius, by far the smaller of our two concerns.

The key idea we will use here is that walker density $D_i$ defined over a sphere centered at the walker $W_i$ position, is roughly -the relation may not be strictly linear- inversely proportional to the average distance from $W_i$ to the other walkers $W_j$.

$$D_i \propto 1 / (\sum Dist(W_i, W_j)/N)$$

Again, N can be eliminated from the equation as we only need to compare proportions, so:

$$D_i \propto 1 / \sum Dist(W_i, W_j)$$

By replacing $Di \propto 1/N_i$ in the previous formula with this new version of $D_i$ we obtain:

$$VR_i \propto R_i * \sum Dist(W_i, W_j)$$

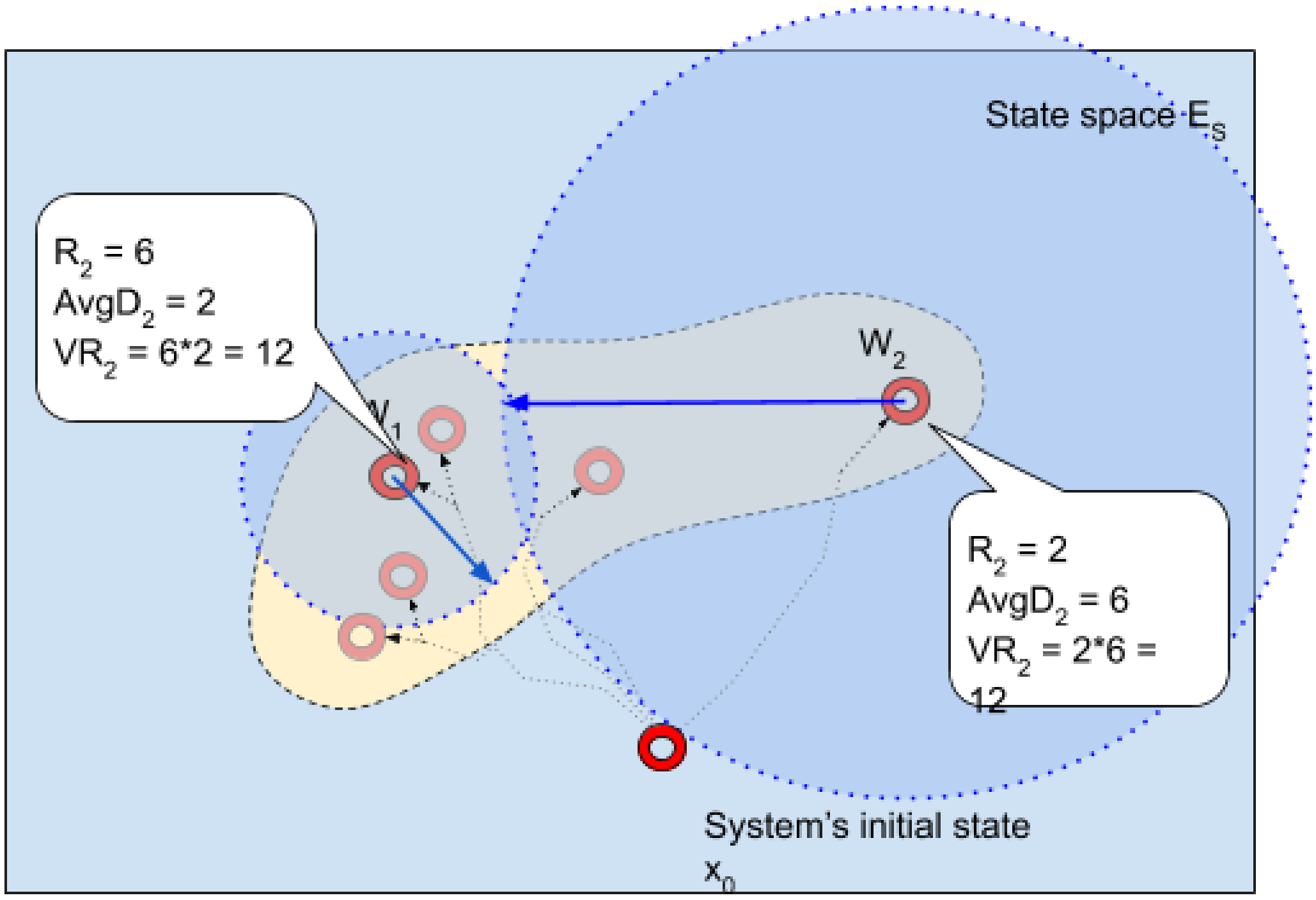

The second simplification we will introduce is quite a dramatic one and it may initially sound like a really bad idea: we will replace the average distance from walker $W_i$ to all the other walkers $W_j$ with just one of those distances, randomly chosen:

$$D_i \propto 1 / \mathrm{Dist}(W_i, W_j) \text{ with j randomly chosen, } j \neq i$$

The resulting virtual reward formulation have a time-complexity of only O(n) while making it to be a highly stochastic function:

$$VR_i \propto R_i / D_i \propto R_i * \mathrm{Dist}(W_i, W_j) \text{ with } j \neq i \text{, randomly chosen.}$$

As our only purpose is to compare virtual rewards, being proportional allows us to safely define:

$$VR_i = R_i * \mathrm{Dist}(W_i, W_j) \text{ with } j \neq i \text{, randomly chosen.}$$

Using this stochastic version of the density actually do a better job than using the standard average distance, and at a much lower computational cost.

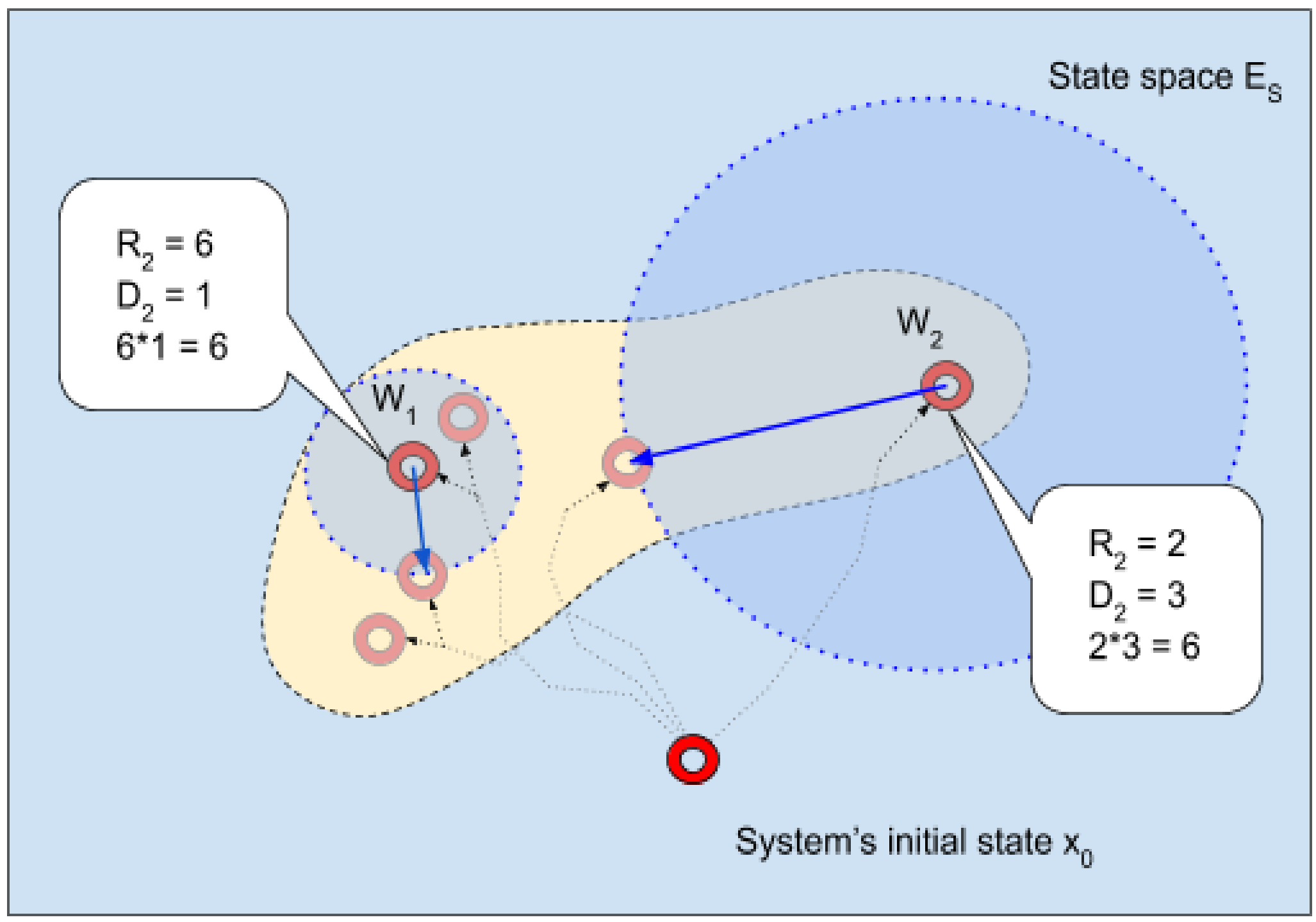

## 4.2.3 - Balancing exploitation and exploration

---

*"Nobody ever figures out what life is all about, and it doesn't matter. Explore the world. Nearly everything is really interesting if you go into it deeply enough."*

---

**Richard Feynman**

Virtual reward is a product of two components: a distance and a reward. They represent the two basic goals of any intelligence: exploration and exploitation.

#### 4.2.3.1 - Relativizing exploitation and exploration

Before we can make any attempt to control this balance, we need to make sure that both components are normalized into a common scale by applying the relativize function introduced in 2.2.3 - Universal reward reshaping to both the rewards $\{R_i\}$ and the distances $\{D_i\}$ between walkers before using them for the Virtual Reward.

$$R_i = (R_i - Mean_R) / Std_dev_R$$
$$R_i = iif(R_i > 0, 1+Ln(R_i), Exp(R_i))$$

$$D_i = (D_i - Mean_D) / Std_dev_D$$
$$D_i = iif(D_i > 0, 1+Ln(D_i), Exp(D_i))$$

#### 4.2.3.2 - Altering the natural balance

A more general formula for the virtual reward can be considered:

$$VR_i = R_i^{\alpha} * Dist^{\beta}(W_i, W_j) \text{ with } j{\neq}i\text{, randomly chosen.}$$

By default both $\alpha$ and $\beta$ are set to 1, meaning that, as far as the distances and rewards are properly scaled, exploitation and exploration will be actually balanced.

The value of $\beta$ is considered fixed at 1 and highly dependent on the metric used, while $\alpha$ is a parameter we can freely change in a standard range from 0 up to 2 or more, actually pushing this balance from equilibrium ($\alpha = \beta$) toward an exploration-only mode ($\alpha = 0$) or toward an aggressive search for reward ($\alpha > \beta$).

#### 4.2.3.3 - The “Common Sense” intelligence

By manually setting $\alpha = 0$, the behaviour changes to a goal-less intelligence, where the decisions are taken in order to increase the number of different reachable futures, regardless of how rewarding they are. The effect is a very clever autopilot that can keep a plane flying around avoiding dangerous paths almost indefinitely.

Please note this “Common sense” mode is nearly equivalent to the idea of maximizing the “empowerment” [6] of the agent as an intrinsic goal. In fact, in this case the agent is maximizing an intrinsic reward represented by the entropy of the available futures after a decision.

Inversely, if $\alpha > \beta$ the agent will be somehow blinded by the reward and will not care much about safety, driving the agent to dangerous situations where the reward is particularly high. If our system can be in death states -our agent can die- then keeping $\alpha$ low is mandatory.

We can then conclude that the value of $\alpha$ roughly represents how safe is it to focus on exploitation in detriment of exploration, or how 'safe' the environment is for the agent.

### 4.2.4 - Probability of cloning

---

*"Often one postulates that a priori, all states are equally probable. This is not true in the world as we see it. This world is not correctly described by the physics which assumes this postulate."*

---

**Richard P. Feynman**

Once we defined a simple yet informative value for the virtual reward of a walker, it is time to get back to the migratory process we outlined before and try to draw a simple method to choose which walkers will be cloning which.

In the process of calculating the virtual reward, a walker must obtain the state of another randomly chosen walker in order to compare positions and obtain a distance. We will also start the cloning process by choosing another random walker in order to compare virtual rewards and obtain a probability of cloning.

Once walker $W_i$ choose a second random walker $W_k$ it will compare virtual rewards and, should he find his to be lower, he could decide to jump to $W_k$ position by cloning its state.

We will define the probability of walker $W_i$ with virtual reward $VR_i$ cloning to $W_k$ state, with virtual reward $VR_k$, as:

- Prob = 1      If $VR_i = 0$
- Prob = 0      If $VR_i > VR_k$
- Prob = $(VR_K - VR_i) / VR_i$      If $VR_i \leq VR_K$

## 4.3 - Pseudo-code

We present a basic pseudo-code implementation to serve as a baseline implementation aimed at simplicity and readability instead of efficiency or scalability.

---

```
// SCALING VALUES:
Relativize(R: array of real)
      M:= R.Mean;
```

```
      S:= R.Std_dev
      IF (S > 0) THEN
            R[i]:= (R[i]-M)/S
            IF (R[i]>0) THEN
                  R[i]:= 1+Ln(R[i])
            ELSE
                  R[i]:= Exp(R[i])

// [0] INITIALIZATION PHASE:
// Define dt for a time horizon Tau, divided on M ticks:
dt:= Tau/M
// Create N walkers with copies of the system's state:
FOR i:= 1 TO N DO BEGIN
      // Walkers start at the system's initial state:
      Walker(i).State:= System.State
      // Walkers take an initial decision:
      Walker(i).Ini_Dec:= random values
      // Walkers simulate their next states:
      Walker(i).State:= Simulation(Walker(i).State, dt)
END
// SCANNING PHASE:
FOR t:= 1 TO M DO BEGIN
      // GET REWARDS AND DISTANCES:
      FOR i:= 1 TO N DO BEGIN
            // Choose a random walker:
            j:= random index from 1 to n, j<>i
            // Read reward and distance:
            R[i]:= Reward(Walker(i).State)
            D[i]:= Distance(Walker(i), Walker(j))
      END
      // NORMALIZE VALUES:
      Relativize(R)
      Relativize(D)
      // CALCULATE VIRTUAL REWARD:
      FOR i:= 1 TO N DO
            Walker(i).VR:= d * Reward(Walker(i).State)
      // UPDATE STATES:
      FOR i:= 1 TO N DO BEGIN
            // Probability of cloning to another walker:
            j:= random index from 1 to n, j<>i
            IF (Walker(i).VR=0) THEN
                  Prob:= 1
            ELSE IF (Walker(j).VR < Walker(i).VR) THEN
                  Prob:= 0
            ELSE
                  Prob:= (Walker(j).VR - Walker(i).VR) / Walker(i).VR
            // Update state by Cloning or Perturbing:
            IF (random < Prob) THEN BEGIN
                  // Cloning:
                  Walker(i).State:= Walker(j).State
                  Walker(i).Ini_Dec:= Walker(j).Ini_Dec
```

```
            END ELSE BEGIN
                  // Perturbing:
                  // Perturbing degrees of freedom:
                  Walker(i).Degrees_of_freedom:= random values
                  // Simulate walker's next state:
                  Walker(i).State:= Simulation(Walker(i).State, dt)
            END
      END
END
// DECIDING PHASE:
Decision:= Sum(Walker(i).Ini_Dec) / N
```

---

Please note that probability of cloning can be >1, feel free to clip it to 1 for formal reasons if this is too uncomfortable for you.

## 4.4 - Classifying the algorithm

The algorithm may actually fit in many of the categories used in the field of algorithmics, making it difficult to properly classify it. Instead, we will name and comment on the categories where it could eventually fit.

Note: This algorithm has an algorithmic time-complexity of O(n), where n is the number of walkers used times the number of ticks we divided the time horizon interval into.

### 4.4.1 - Monte Carlo Planning algorithm

---

*"We shall not cease from exploration*
*And the end of all our exploring*
*Will be to arrive where we started*
*And know the place for the first time."*

---

***T.S. Eliot, Four Quartets***

One of the most evident classification of the algorithm is as a Planning algorithm, similar to Monte Carlo Tree Search (MCTS), where the different futures a system can visit are scanned up to some depth level (a time horizon) while forcing the less appealing branches to be rejected in the process in order to focus on the most promising ones.

The main differences between Fractal Monte Carlo (FMC) and the many MCTS variants [4] are:

1. MCTS is usually used in games where two players fight one against the other. FMC is defined for one player games, but can be easily adapted to other scenarios.
2. MCTS only deal with discrete decision spaces. FMC deals with both cases.
3. MCTS build the decision tree one path at a time. FMC builds a big number of branches simultaneously with a swarm of walkers interacting one with each other.

4. MCTS needs to keep the whole tree structure. FMC only uses the leaves of the tree.
5. MCTS resources grow exponentially with scanning depth. In FMC the CPU resources grows linearly and memory resources doesn't grow with depth.

### 4.4.2 - Cellular automaton

The concept of "walker" as an entity that uses an internal programing to autonomously decide on its evolution, is easily assimilable to the concept of a cellular automaton, except for the fact that, in our case, the system and the available actions don't need to be discrete.

In fact, from the standpoint of a walker, the algorithm is just a repetition of a series of small steps that, when simultaneously performed by a number of walkers, automatically generates an intelligent decision on the agent:

1. Choose a random walker and get its state.
2. Calculate Virtual Reward.
3. Choose a second random walker and get its state.
4. Compare Virtual Rewards and decide about cloning.
5. Clone the second walker's state if decided.
6. Randomly perturb your state.

### 4.4.3 - Swarm algorithm

FMC is also a swarm intelligence algorithm where the collective behaviour of a pool of decentralized, self-organized agents, solely determines the decision to be made.

Please note that, although in the pseudo-code the walkers are moved and cloned in a synchronized way for the sake of clarity, it is not mandatory at all, walkers could communicate via asynchronous messages and evolve in an asynchronous and isolated way, making it extremely easy to parallelize or distribute the algorithm in order to scale it. This, along with the linear time-complexity, makes the algorithm highly scalable.

### 4.4.4 - Evolutionary algorithm

FMC is a population-based evolutionary algorithm where walkers are divided into population groups (based on their initial actions) that compete for success by cloning the best fits (highest virtual reward) overwriting the worst fits, and then mutating them by randomly changing their states on the simulating phase.

### 4.4.5 - Entropic algorithm

There is not a real category of "entropic" algorithms, but it should. An algorithm is entropic when it is driven by the maximization/minimization of some kind of entropy, cross-entropy, or divergence.

Gradient descent optimization of a loss function like cross-entropy or the divergence of two distributions, as used in deep learning, is a clear example of an entropic algorithm.

FMC is an entropic algorithm, as it is based on minimizing the divergence -or equivalently maximizing a cross-entropy- of two distributions: reward and walker probability distributions.

### 4.4.6 - Fractal algorithm

In a Monte Carlo Tree Search, where actions are discrete, the graph of the states we visit in the search process is a tree. FMC generate the same kind of trees in the same conditions.

When the decision and the state spaces are both continuous, then the distances between walkers and the time step dt we use for the simulation can we made as small as we want making the resulting tree to be formed by smaller and smaller pieces.

In the limit when both the number of ticks used to divide the time horizon and the number of walkers tend to infinity, the graph morphs from a finite tree to a fractal tree.

# 5 - Experiments

We have run several experiments in order to show the algorithm strong points and to compare FMC with other similar algorithms in both the discrete and continuous cases.

## 5.1 - Discrete case: Atari 2600 games

One of the most widely accepted benchmarks for AI methods are the Atari 2600 games from OpenAI Gym. They all have a discrete decision space of six on/off buttons and well defined scorings allowing for fair comparison between algorithms.

To test FMC against other algorithms, we will be using a set of 50 Atari games commonly used in all the planning algorithm literature, even if some other articles -related to learning algorithms- may add some 5 to 7 extra games.

### 5.1.1 - Results

FMC will be compared against four relevant baselines:

1. Standar Human: score obtained by a professional game tester after 2 hours of training.
2. Human World Record [A2]: maximum score ever reported from a human being.
3. Planning algorithms [P1-7]: MCTS UCT, IW(1), p-IW(1), 2BSF, BrFS, etc.
4. Learning algorithms [L1-9]: DQN, A3C, C51 DQN, Dueling, NoisyNet-DQN, NoisyNet-A3C, NoisyNet-Dueling, HyperNEAT, etc.

Being FMC a planning algorithm, the only "apples-to-apples" comparison is against State of the Art (SoTA) planning algorithms. The rest of the comparisons, especially against learning algorithms, are actually "apples-to-bananas" as they have no access to a simulation of the model to sample future states from, instead they learn from perfect memories of past experiences.

That said, our experiments showed that:

| | Wins | % |
|---|---|---|
| FMC vs Standard Human | 49 / 50 | 98% |
| FMC vs Human World Record | 32 / 50 | 64% |
| FMC vs Best Planning SoTA | 50 / 50 | 100% |
| FMC vs Best Learning SoTA | 45 / 50 | 90% |
| FMC "solved" the game | 32 / 50 | 64% |
| Solved due to the 1M bug | 16 / 50 | 32% |

We will consider a game as being "solved" when:

a) Reached the ending score (24 for Tennis and Double dunk, 100 for Boxing or 21 for Pong).
b) Reached a hard-coded maximum score (999,500 in Asterix).
c) Reached a bug-induced limit score (many games reset score above 999,999 like Ms Pacman, Chopper command, Daemon attack, Frostbite or Gopher, while Wizard of Wor limit is 99,999).
d) Reached "immortality", being able to play indefinitely until the game was manually halted (Atlantis and Asteroids were halted once above 10M).
e) Scored above all known records, specially the actual human world record.

Please note that, in the case of bug-limited games, the reported human world records cannot be reached by any program, as the human records can include an additional 'units of million' digit that did never showed up on the game screen, nor in the internal score accessed via APIs.

In the following high score table, solved games use gray background and are labeled with the reason why they were considered as solved.

| Game | FMC Score | Standard Human | Human Record | Planning SoTA | Learning SoTA |
|---|---|---|---|---|---|
| ***Alien (e)*** | **479,940** | 7,128 | 251,916 | 38,951 | 7,967 |
| ***Amidar*** | 5,779 | 1,720 | **155,339** | 3,122 | 4,058 |
| ***Assault (e)*** | **14,472** | 1,496 | 8,647 | 1,970 | 11,734 |
| ***Asterix (b)*** | **999,500** (1) | 8,503 | 335,500 | 319,667 | 406,211 |
| ***Asteroids (d)*** | **12,575,000** | 47,389 | 10,004,100 | 68,345 | 167,159 |
| ***Atlantis (d)*** | **10,000,100** | 29,028 | 7,352,737 | 198,510 | 2,872,645 |
| ***Bank heist*** | 3,139 | 753 | **199,978** | 1,171 | 1,496 |
| ***Battle zone (b)*** | **999,000** | 37,800 | 863,000 | 330,880 | 53,742 |
| ***Beam rider (c)*** | **999,999** | 16,926 | **999,999** | 12,243 | 21,077 |
| ***Berzerk*** | **17,610** | 2,630 | **1,057,940** | 2,096 | 2,500 |
| ***Bowling*** | **180** | 161 | **300** | 69 | 136 |
| ***Boxing (a)*** | **100** | 12 | **100** | **100** | **100** |
| ***Breakout (e)*** | **864** | 31.8 | 752 | 772 | 748 |
| ***Centipede (d)*** | **1,351,000** | 12,017 | 1,301,709 | 193,799 | 25,275 |
| ***Chopper command (c)*** | **999,900** | 9,882 | **999,900** | 34,097 | 15,600 |
| ***Crazy climber (d)*** | **2,254,100** | 35,829 | 447,000 | 141,840 | 179,877 |
| ***Demon attack (c)*** | **999,970** | 3,401 | **1,556,345** (2) | 34,405 | 130,955 |

| ***Double dunk (a)*** | **24** | -15.5 | **24** | **24** | **24** |
|---|---|---|---|---|---|
| ***Enduro (e)*** | **5,279** | 860 | 3,618 | 788 | 3,454 |
| ***Fishing derby*** | 63 | 6 | **71** | 42 | 59 |
| ***Freeway*** | 33 | 30 | **34** | 32 | **34** |
| ***Frostbite (c)*** | **999,960** | 4,335 | 552,590 | 6,427 | 4,442 |
| ***Gopher (c)*** | **999,980** | 2,412 | 120,000 | 26,297 | 41,138 |
| ***Gravitar*** | 14,050 | 1,693 | **1,673,950** | 6,520 | 2,308 |
| ***Hero*** | 43,255 | 30,826 | **1,000,000** | 15,280 | 105,929 |
| ***Ice hockey (e)*** | **64** | 1 | 36 | 62 | 11 |
| ***James bond (e)*** | **152,950** | 407 | 45,550 | 23,070 | 6,963 |
| ***Kangaroo*** | 10,800 | 3,035 | **1,436,500** | 8,760 | 15,470 |
| ***Krull*** | 426,534 | 2,666 | **1,006,680** | 15,788 | 35,024 |
| ***Kung fu master*** | 172,600 | 22,736 | **1,000,000** | 86,290 | 79,676 |
| ***Montezuma revenge*** | 5,600 | 4,753 | **1,219,200** | 500 | 4,740 |
| ***Ms Pacman (c)*** | **999,990** | 15,693 | 290,090 | 30,785 | 5,913 |
| ***Name this game (e)*** | **53,010** | 8,049 | 25,220 | 15,410 | 12,542 |
| ***Pong (a)*** | **21** | 15 | **21** | **21** | **21** |
| ***Private eye*** | 41,760 | 69,571 | **103,100** | 2,544 | 40,908 |
| ***Q-Bert (c)*** | **999,975** | 13,455 | **2,400,000** [2] | 44,876 | 27,543 |
| ***River raid*** | 18,510 | 17,118 | **194,940** | 15,410 | 24,568 |
| ***Road runner (c)*** | **999,900** | 7,845 | **2,038,100** [2] | 120,923 | 367,023 |
| ***Robo tank (e)*** | **94** | 12 | 74 | 75 | 65 |
| ***Seaquest (c)*** | **999,999** | 42,055 | 527,160 | 35,009 | 266,434 |
| ***Space invaders*** | 17,970 | 1,669 | **621,535** | 3,974 | 7,227 |
| ***Star gunner (c)*** | **999,800** | 10,250 | 77,400 | 14,193 | 84,490 |
| ***Tennis (a)*** | **24** | -8 | **24** | **24** | 23 |
| ***Time pilot (e)*** | **90,000** | 5,925 | 66,500 | 65,213 | 18,501 |
| ***Tutankham*** | 342 | 168 | **3,493** | 226 | 288 |
| ***Up n down (c)*** | **999,999** | 11,693 | 168,830 | 120,200 | 155,049 |
| ***Venture*** | 1,500 | 1,188 | **31,900** | 1,200 | 1,520 |
| ***Video pinball (c)*** | **999,999** | 17,668 | **91,862,206** (2) | 471,859 | **999,999** |
| ***Wizard of Wor (c)*** | **99,900** | 4,756 | **233,700** (2) | 161,640 | 16,143 |
| ***Zaxxon*** | 92,100 | 9,173 | **100,000** | 39,687 | 18,057 |

(1) Records obtained by @spirosbax.
(2) Human record is higher than the internal score limit (1M bug).

## 5.1.2 - Sampling efficiency

Planning algorithms work by sampling the model many times in order to know about the family of future states we can reach from an initial state after a number of steps. In practice, one or several copies of the Atari environment are used as perfect simulators to build those paths step by step by chaining one state with its predicted next state after taking one action.

Efficiency is usually defined as "samples per action", the mean number of samples of the model we needed to query before an action was taken.

All the benchmarked planning algorithms use a minimum of 150,000 samples per action while FMC, being specially cheap on sampling, used on average 359 times fewer samples per action.

## 5.1.3 - Implementation details

The Atari games experiment was implemented in python following the ideas in this document.

### 5.1.3.1 - Github repository

You can find the full python code for the Atari experiment at:

https://github.com/FragileTheory/FractalAI

### 5.1.3.2 - Using two FMC flavours

Two different implementations of the FMC algorithm were used for the tests. You can find both version in the GitHub repository.

In the first implementation -the "step-by-step" standard FMC version- we follow the ideas presented in the pseudo-code, where the agent makes each decision by building a totally new cone after each step, starting at the actual position of the agent, and using a fixed number of walkers and tree depth (time horizon).

In a second implementation -the "swarm wave" FMC version- the agent build a single cone, starting in its initial state without limiting to any time horizon. We make this cone to grow until one of the paths reaches a desired limit score. Once this goal is met, the algorithm stops and a video of the highest scoring path is shown, while a dataset of thousands of high quality episodes is stored ready to be used to train a DNN.

#### 5.1.3.3 - RAM vs Images

A game can send the observation of its actual state in two flavours:

1. The screen image as an RGB array.
2. The internal RAM state as a byte array.

FMC base its decisions on the entropic properties of the swarm of walker states. Being this an intrinsic goal, FMC can be equally applied to RAM or image-based observations with very similar results.

As most of the methods we will be comparing with can only be played on image-based games, we can only fairly compare IMG vs RAM observations solved with FMC using the same parameters (fixed_steps=5, time_limit=15, Max_walkers=30, Max_samples=300). Even with those low values of 300 samples per step, FMC is able to beat state of the art (SoTA) algorithms and totally solve some of the games:

| | FMC on IMG | FMC on RAM | RAM vs IMG |
|---|---|---|---|
| **atlantis** | 145,000 | 139,500 | 96.21% |
| **bank heist** | 160 | 280 | 175.00% |
| **boxing** | 100 | 100 | 100.00% |
| **centipede** | Immortal (1) | immortal | 100.00% |
| **ice hockey** | 16 | 33 | 206.25% |
| **ms pacman** | 23,980 | 29,410 | 122.64% |
| **qbert** | 17,950 | 22,500 | 125.35% |
| **video pinball** | 1,273,011 (2) | 1,999,999 | 366.29% |
| | | | **161.47%** |

(1) Centipede was halted scoring +600K and having 7 extra lives.
(2) Video pinball score resets to 0 after 999,999. RAM managed to stay at 999,999 for 4 minutes, while IMG couldn't avoid it.

It may seem counter-intuitive that FMC scored better on RAM without seeing the screen images, but actually it makes quite sense: RAM is the actual state of the game, while images are partial observations that contains a lot of noise (a single bit changed on RAM can cause a firework on screen), so the distance between RAM dumps is far more informative than knowing the distance between the corresponding screen images.

## 5.2 - Continuous case: Flying rocket

In this experiment we control a 2D rocket flying inside a closed environment. The rocket has two continuous degrees of freedom corresponding to the trust levels for the main rocket and a secondary one used for rotating.

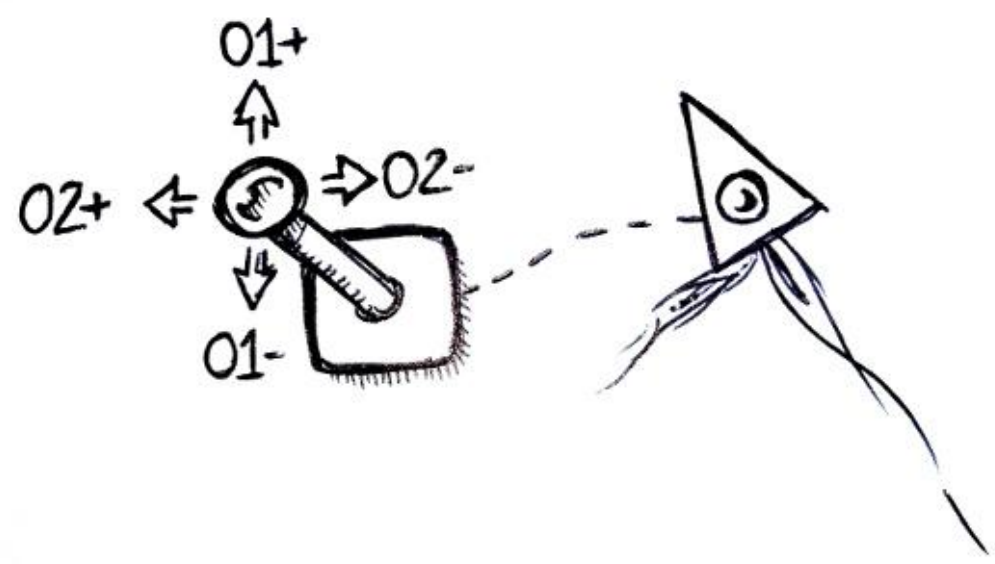

At each step we define the force applied to each of the degrees of freedom, so our decision space is a continuous 2D space. In these cases we will not be choosing an action from a list, instead we will be deciding on a force as a 2D vector.

The rocket is a very interesting toy-system, it is very dynamic as equilibrium is never reached and making fast decisions is critical. When FMC is used in a continuous decision system like this, it performs even better than it did in the discrete case. Being the decision an average of many decisions, it is more naturally defined in the continuous case than in the discrete one.

### 5.2.1 - Flying a chaotic attractor

> *"Chaos was the law of nature; Order was the dream of man.*
>
> **Henry Adams**

As flying a rocket was not a big problem for fractal AI, we designed a special environment where the goals were almost impossible to get.

First, a hock was attached to the rocket using a rubber band, forming a chaotic oscillator where the final position of the hook is highly sensitive to small changes in the initial conditions, making it extremely difficult to sample the state space and find the low probability paths that define the right decision.

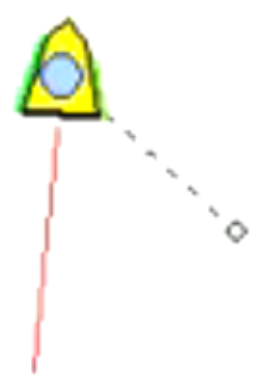

Secondly, we will define a ridiculous difficult goal for the system: drive the rocket in such a way that the chaotic hook picks a falling rock, take it into a distant deploy zone (crossed area), release it and wait until it leaves the dotted circle, and repeat it as many times as you can.

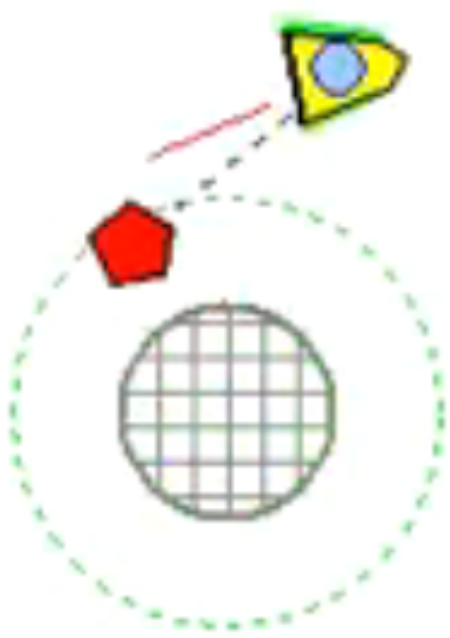

To define this 'hook goal' we just have to convert it into a reward function.

1. For every walker, the hook defines its target, being it the nearest rock if the hook is empty, and the deploy area when it is holding a rock.
2. For every walker, the hook calculates the distance D to the target.
3. Now the walker compares it with the same distance calculated at the initial state (the agent position).
4. If distance is 0, you did it! Hook Reward = 100
5. If distance is bigger: Hook Reward = 10e-10
6. If distance is 20% smaller: Hook Reward = 0.2

Only this reward function was added, the rest of the code was just the standard FMC code and a homebrew physics simulator of the system.

### 5.2.2 - Results

Fractal AI was able to solve this problem using only 300 walkers (paths) and 200 samples per walker (a time horizon of 2 seconds at steps of 0.1 second) for a total of 60,000 samples per decision.

The agent was able to chain several goals in a row, and also could recover the rock when it fell down to the ground, totally solving the task with extremely low computational resources.

Video 1: Solving the task (https://youtu.be/HLbThk624jI)

This experiment also allows us to watch the fractal paths generated in the process.

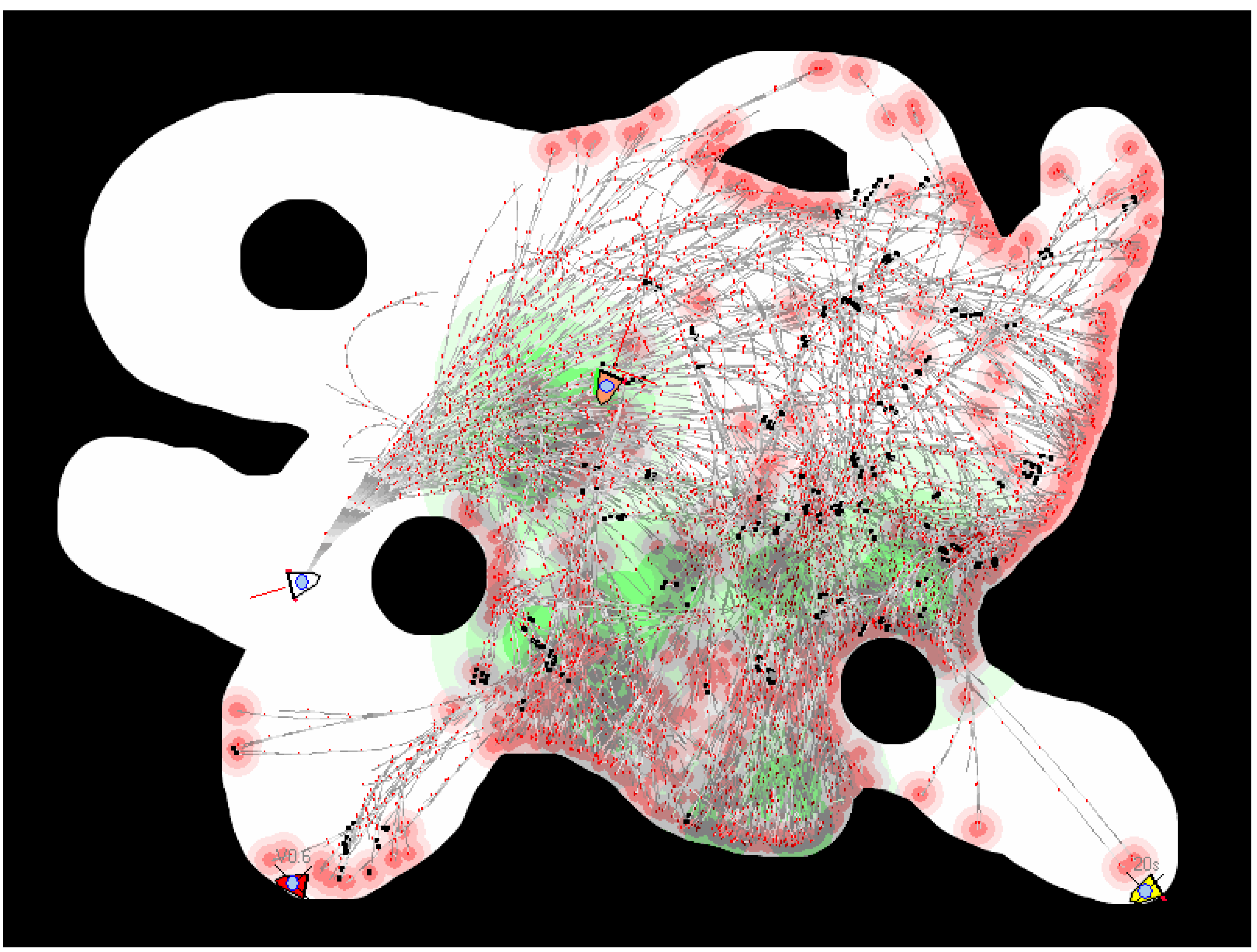

Video 2: [Visualizing the decision process](https://youtu.be/cyibNzyU4ug) (https://youtu.be/cyibNzyU4ug )

## 5.2.3 - Implementation details

This experiment used an early implementation of the algorithm that was 99% the same as in the pseudo-code. It is coded in object-pascal (delphi 7) and includes a complete ad-hoc physical simulator.

The main differences with the general case came from the definition of distance between two states and the reward function used.

### 5.2.3.1 - Reward

The reward was the composition of two goals: keep alive -that correspond to the rocket health level (the simulator used the energy of the collisions to take health from the agent)- and the already commented 'hook reward', so the actual reward was defined as:

Reward = Health_Level * Hook_Reward

### 5.2.3.2 - Distance

Any informative distance would have made the trick, but as this was a very physical example, we decided to focus only on the position and momentum of the rocket to define its position, thus the distance used was:

Dist(A, B) = Sqrt( (A.x-B.x)^2 + (A.vx-B.vx)^2 + (A.y-B.y)^2 + (A.vy-B.vy)^2 )

Adding the rest of the coordinates to the distance formula resulted in a lower performance. In general, using a distance that makes sense in the problem helps. For the general case, using an embedding of the state can reduce its dimensionality and, as in any other method, helps to improve efficiency.

# 6 - Research topics

We will briefly comment on some of the research topic we find worth exploring.

## 6.1 - Distributed computing

The algorithm time complexity is O(n) where n = num. walkers x num. ticks, but walkers work almost independently -except for some inter-walker communication- and even asynchronously, so a parallel implementation of the algorithm, assigning one core per walker can, in principle, lower the time complexity near to O(num. ticks).

Thus, adding more CPU can scale up the algorithm almost linearly but, eventually, the inter-processes communication overhead of sending system states from one walker to another will impose a practical limit dependent, in practical terms, on the size of the system state.

Reached this point, a second distributed strategy is launching several FMC processes -workers- in a distributed environment, each one using a smaller number of walkers to output a decision vector that will be averaged by the agent in order to make its decision. This 'clustered' decision is almost as reliable as using the sum of all walkers in a single fractal.

By adding the capability to reshuffle the states of all the walkers among the cluster of workers every m steps, you can continuously change from totally isolated workers (m = number of total ticks, no communication overhead) or totally connected workers (m = 1, overhead depending on state size and simulation time).

Please note that the repository contains a parallel version that uses as many cores as you wish (and have) but it is not a truly distributed version as it runs still on a single PC.

## 6.2 - Adding learning capabilities

---

*"One remembers the past to imagine the future"*

---

**Daniel L. Schacter**

As we noted before, one of the limitations of the algorithm was dealing only with the forward-thinking process, so one natural research topic is mixing forward and backward-thinking in a single process.

The presented algorithm scans all available actions at any given state with a uniform probability distribution, it doesn't have any 'a priori' preference, so the walks produced are truly random.

FMC algorithm could be used to feed examples of correct decisions -rollouts- to a neural network that, in turn, would train itself to predict the intelligent decision -a probability distribution over the actions- as a function of the state.

We can then use this as the 'a priori' distribution: when a walkers needs to randomly choose an action before simulating a delta of time, instead of choosing a random action, we can now sample it from this distribution.

A mechanism like this could give the walkers a natural tendency to repeat actions that worked well in the past on similar situations. Usually it means that, with the same number of walkers, you can get better decisions or, inversely, that you can safely lower the number of walkers needed to decide on situations that are familiar to the agent.

In the extreme case where the neural network can be considered well trained, you can completely disable walkers and decide by choosing the most probable action in the NN output.

By comparing the decision suggested by such a memory system with the one generated by the FMC algorithm -for the same initial state- we can estimate how accurate our a priori distribution is, enabling us to dynamically adjust the 'credibility' of this distribution.

To get a simplified sketch of the idea we could:

1) Define a function of the state that outputs a distribution over the actions: N(State) = ($N_i$).
2) When the algorithm ends, we were already getting a distribution ($P_i$) over the available actions (the proportion of walkers associated with each action).
3) We can know how similar they are: Credibility = $D_H(P_i, N_i) / D_H(P_i, \text{Uniform})$.
4) In the next iteration we can reduce the number of walkers to: Max_Walkers * (1-Credibility)
5) In this next iteration, when walkers are building the random walks, they first get the 'a priori' distribution ($N_i$) associated to its actual state, and mix it with the uniform distribution ($U_i$)=(1/N): ($X_i$) = ($N_i$)*Credibility + ($U_i$)*(1-Credibility).

This mechanism opens the possibility of mixing FMC with any general learning method -like neural networks- in such a way that it can detect when the memory-based backward-thinking "fast" decisions are good enough to replace the more expensive -in terms of CPU usage- forward-thinking decisions, allowing us to:

1. Dynamically reduce the number of walkers used as the credibility of the distribution gets higher to speed up the process.
2. Get better results over time using the same number of walkers.
3. Generate a stand-alone fast policy (a standard neural network) to make decisions in absence of walkers (backward-thinking only).

### 6.2.1 - Using a DQN for learning

DQN is a model of deep learning designed to learn the probability of the actions -expressed as reward expectation- as a function of the system state and as such it is a perfect match for FMC.

In one hand, FMC is able to generate good quality game rollouts -sequences of pairs state-action from previous games without the need of a priori density of probabilities for actions, so it is able to feed the learning process of the DQN with meaningful game sequences instead of random played ones, boosting the learning performance to some degree.

On the other hand, once the DQN has learned from a dataset of initial rollouts, FMC can use it output as a priori for randomly choosing an action at each walker step, making the FMC to scan more deeply those actions suggested by the DQN and thus, making FMC more efficient and capable as the DQN learns to make better predictions.

This kind of combination is not new in the literature [7] but replacing UCT (a state-of-the-art implementation of MCTS) with FMC could improve the efficiency of the hybrid. A fair comparison between both methods is presented in the experiments section showing that FMC can outperform UCT using about 0.01% to 0.1% of the samples per step.

## 6.3 - Common Sense Assisted Control

Imagine a drone driven with FMC, following the only goal of not crashing into anything. Now add a remote control that, when pushed forward, send an a priori probability over the actions where going forward is much more probable that all the other actions, in the same way the DQN would be doing.

The FMC will try to follow this direction, but only if this doesn't go against its first goal of not crashing into anything. The resulting drone will follow the control orders without crashing, allowing to fly it on difficult scenarios with easy and safety.

## 6.4 - Consciousness

---

*"In any decision for action, when you have to make up your mind what to do, there is always a 'should' involved, and this cannot be worked out from, 'If I do this, what will happen?' alone."*

---

**Richard P. Feynman**

When the reward function is a composition of several goals $\{G_i\}$ we can assign a relative importance $\{K_i\}$ for each goal, having $K_i \geq 0$ and $\Sigma(K_i)=1$, so our reward function would looks like this:

$$\text{Reward}(X) = \Pi(G_i(X)^{K_i})$$

We can consider the vector $\{K_i\}$ as being the "mental state" -as opposed to the physical state- of the agent. Any mechanism that could automatically adjust those coefficients in order to make better decisions can be considered as a conscious mechanism.

If we consider those $\{K_i\}$ as being a second-level agent state, we can use them as both state component and degrees of freedom, and apply the same FMC algorithm to intelligently adjust them... as far as we can define a sensible reward associated with a mental state $\{K_i\}$.

## 6.5 - Real-time decision-making

*"Life is the continuous adjustment of internal relations to external relations."*

**Herbert Spencer**

The need in the present implementation of resetting all the walkers to the new agent position after a step is made, force us to erase valuable information, making the samples per step to be higher than it could be.

A slightly different implementation could lead to a continuous decision algorithm where a new decision is generated for each tick of the algorithm -as opposed to each step of the agent- in a totally continuous process, allowing to sample the decision at any time the agent needs it.

In this new implementation, each continuous decision would come with a measure of the average delay between the agent time when the walkers started its path and the actual agent time when the decision is used. A faster CPU would allow the algorithm to have a smaller delay, use more walkers and paths, or think at a longer time horizon. A new parameter would be needed to limit this delay below a desired value.

## 6.6 - Universality pattern

*"Find beauty not only in the thing itself, but also in the pattern of the shadows, the light and dark which that thing provides"*

**Junichiro Tanizaki**

It would be interesting to connect the distances between walkers at any moment with the Universality pattern found in the eigenvalues of random matrices. This pattern seems to be universal to any system where the parts are heavily correlated. In the pool of walkers it is the case, so if this algorithm is some form of universal complex system solver, it makes sense to try to connect both ideas.

# 7 - Conclusions

The theory of intelligence introduced allowed us to build a very efficient agent that, in the discrete decision case, outperforms actual implementations of MCTS and other planning algorithms in two or three orders of magnitude. This is done by just inspecting the tree of decisions using entropy-based principles while boosting exploration to achieve a balance in the exploitation vs exploration tradeoff.

The algorithm can also be applied to continuous decision spaces, where it proved to be highly efficiently, introducing Monte Carlo methods into a new spectrum of possible uses like driving vehicles or controlling robots.

The algorithm naturally allows working with neural networks in a close symbiosis where the NN learns from the actions taken by the intelligence to produce a good prior distribution over the actions, that is then used by the planning algorithm to get better results over time, that again are fed into the learning process of the NN, closing a virtuous cycle of improvements. By doing so, actual reinforced learning methods could be reshaped into a simpler and more efficient form of supervised learning.

## Atari experiment references:

### Planning algorithms:

## Learning algorithms: